\documentclass{article}

\usepackage[nonatbib, preprint]{neurips_2023}
\usepackage{microtype}
\usepackage{graphicx}
\usepackage{subfigure}

\usepackage[utf8]{inputenc} % allow utf-8 input
\usepackage[T1]{fontenc}    % use 8-bit T1 fonts
\usepackage{hyperref}       % hyperlinks
\usepackage{url}            % simple URL typesetting
\usepackage{booktabs}       % professional-quality tables
\usepackage{amsfonts}       % blackboard math symbols
\usepackage{nicefrac}       % compact symbols for 1/2, etc.
\usepackage{microtype}      % microtypography
\usepackage{xcolor}         % colors
\usepackage{MnSymbol}   
\usepackage{amsmath}
\usepackage{booktabs}
\usepackage{algorithm}
\usepackage{algorithmic}
\usepackage{lineno}
\usepackage{bm}
\usepackage{bbm}
\usepackage{float}
\usepackage{multirow}
\usepackage{makecell}
\usepackage{tabularx}
\usepackage{stmaryrd}
\usepackage{lipsum}
\usepackage{subfigure}
\usepackage{bbold}
\usepackage{amsthm}
\usepackage{hyperref}
\usepackage{newfloat}
\usepackage{listings}
\usepackage{wrapfig}
\usepackage{tikz}

\hypersetup{colorlinks=true,
citecolor=green,
linkcolor=red,
urlcolor=blue
}

\theoremstyle{plain}
\newtheorem{theorem}{Theorem}[section]

\newtheorem{corollary}[theorem]{Corollary}
\theoremstyle{definition}
\newtheorem{definition}[theorem]{Definition}

\theoremstyle{remark}

\title{Introducing Expertise Logic into Graph Representation Learning from A Causal Perspective}

\author{%
  Hang Gao, Jiangmeng Li \thanks{Corresponding Author}\ ,  Wenwen Qiang, Linyu Si, Xingzhe Su\\
  Institute of Software Chinese Academy of Sciences\\
  No. 4, South 4th Street, Zhongguancun, Haidian District, Beijing, China\\
  University of Chinese Academy of Sciences\\
  No. 19 (A), Yuquan Road, Shijingshan District, Beijing, China \\
  \texttt{\{gaohang, jiangmeng2019, qiangwenwen, lingyu, xingzhe2018\}@iscas.ac.cn} \\
  \And
  Fengge Wu, Changwen Zheng \\
  Institute of Software Chinese Academy of Sciences\\
  No. 4, South 4th Street, Zhongguancun, Haidian District, Beijing, China\\
  \texttt{ \{fengge,changwen\}@iscas.ac.cn} \\
    \And
  Fuchun Sun \\
  Tsinghua University \\
  No. 30, Shuangqing Road, Haidian District, Beijing\\
  \texttt{fcsun@mail.tsinghua.edu.cn} \\
}

\begin{document}

\maketitle

\begin{abstract}

Benefiting from the injection of human prior knowledge, graphs, as derived discrete data, are semantically dense so that models can efficiently learn the semantic information from such data. Accordingly, graph neural networks (GNNs) indeed achieve impressive success in various fields. Revisiting the GNN learning paradigms, we discover that the relationship between human expertise and the knowledge modeled by GNNs still confuses researchers. To this end, we introduce motivating experiments and derive an empirical observation that the GNNs gradually learn human expertise in general domains. By further observing the ramifications of introducing expertise logic into graph representation learning, we conclude that leading the GNNs to learn human expertise can improve the model performance. Hence, we propose a novel graph representation learning method to incorporate human expert knowledge into GNN models. The proposed method ensures that the GNN model can not only acquire the expertise held by human experts but also engage in end-to-end learning from datasets. Plentiful experiments on the crafted and real-world domains support the consistent effectiveness of the proposed method.

%By exploring the intrinsic mechanism behind such observations, we elaborate the Structural Causal Model for the graph representation learning paradigm. Following the theoretical guidance, we innovatively introduce the auxiliary causal logic learning paradigm to improve the model to learn the expertise logic causally related to the graph representation learning task. In practice, the counterfactual technique is further performed to tackle the insufficient training issue during optimization. Plentiful experiments on the crafted and real-world domains support the consistent effectiveness of the proposed method.

\end{abstract}

\section{Introduction} \label{sec:intro}

As the improvement of machine computing capacity hits a developmental bottleneck, barely increasing the network parameters and the number of input data becomes challenging to boost the model performance. Therefore, researchers explore empowering the model's capacity by leveraging human prior knowledge to guide optimization. An outstanding practical attempt is converting the natural data structures, e.g., images and videos, into derived discrete data structures, e.g., language and \textit{graph}. Benefiting from the injection of human prior knowledge, derived discrete data structures are generally semantically dense. Thus, exploring the semantic information from such derived data can effectively improve the model performance.

By revisiting the learning paradigms of benchmark methods, we observe that the improvements according to GNNs essentially lie in two aspects: 1) the message passing process, which aggregates features from a node's neighbors, e.g., GGAE \cite{DBLP:journals/tnn/LiWFNZ22}, $p$-Laplacian GNN \cite{DBLP:conf/icml/FuZB22}, PG-GNN \cite{DBLP:conf/icml/HuangWLH22}, GloGNN \cite{DBLP:conf/icml/LiZCSLLQ22}; 2) the readout function, which projects the node representations into the graph representation, e.g., Adaptive Structure Aware Pooling \cite{DBLP:conf/aaai/RanjanST20}, Top-k Pool \cite{DBLP:conf/icml/GaoJ19}, SAG Pool \cite{DBLP:conf/icml/LeeLK19}, etc. State-of-the-art approaches elaborate GNN structures to learn discriminative representations from graphs. However, there exist several innate questions challenging current researchers who explore GNN-based learning paradigms: what kind of knowledge do GNNs capture from graphs and is such knowledge in accordance with human expertise?

\begin{figure} \centering
\subfigure[Visualization of a conventional GNN approach's learning degree of the expertise logic during optimization on benchmark datasets. We specifically measure the cosine similarity between the logits of the expert model and the GNN model. Generally, the results show that the similarity has an increasing trend with training.] { \label{logicgnn}
\includegraphics[width=0.35\columnwidth]{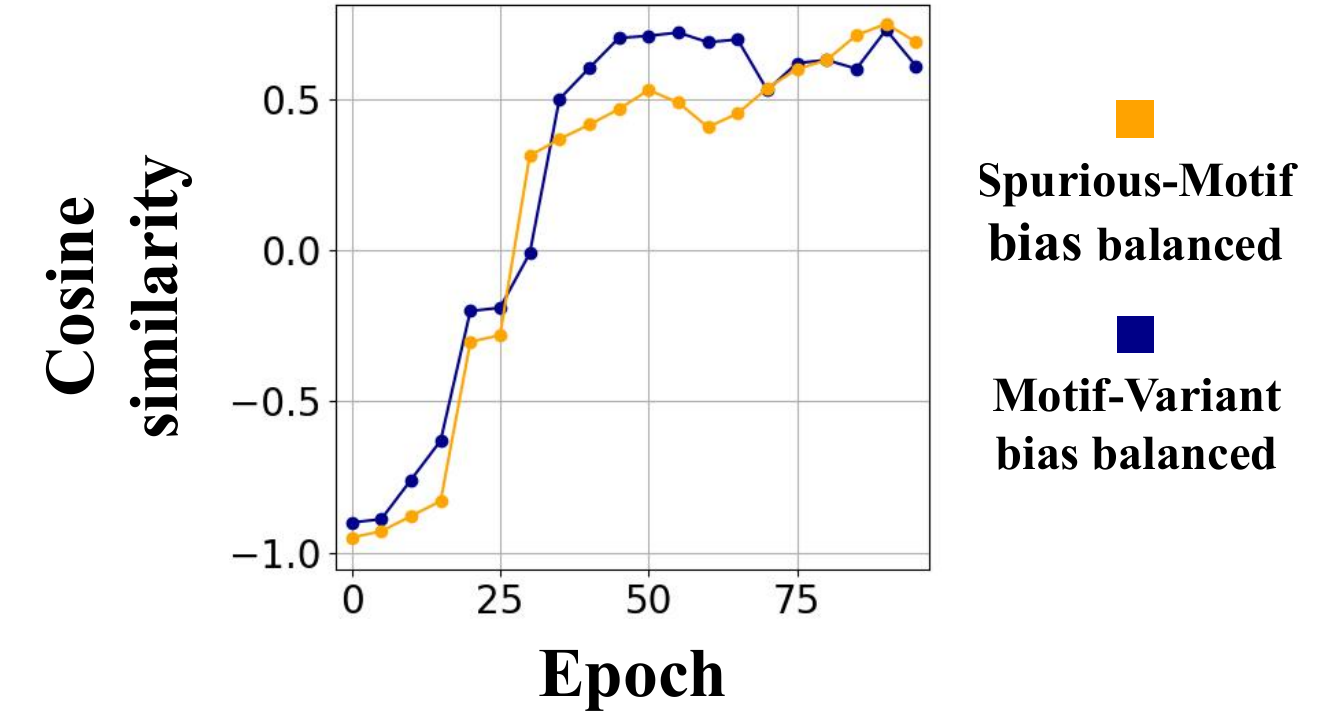}
}\hspace{3mm}
\subfigure[Comparisons between the baseline, i.e., Empirical Risk Minimization (ERM), and the proposed method on various datasets. The left plot demonstrates the KL divergence between the features derived by the expert model and the candidate model. The right plot collects the classification performance of the compared methods. The empirical observations jointly support that our proposed method can better learn the expertise logic and consistently outperform the baseline.] { \label{gnnclglcomp}
\includegraphics[width=0.58\columnwidth]{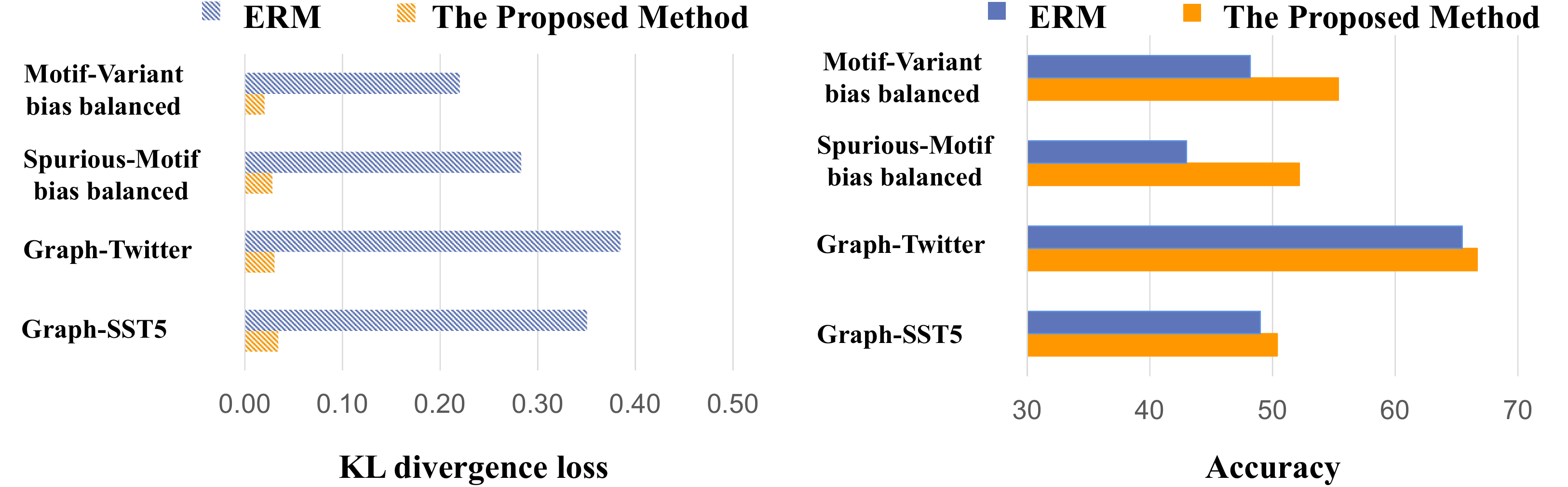}
}
\vskip -0.1in
\caption{ The motivating explorations. }
\vskip -0.2in
\label{fig}
\end{figure}

Inspired by this motivation, we intuitively introduce the experiments to explore the knowledge captured by GNNs. As demonstrated in Figure \ref{logicgnn}, the expertise logic contained by the expert model is continuously learned by the purely Connectionism-based data-driven GNN model, and we thus derive the intuition that GNNs indeed gradually capture the knowledge that is homogeneous to the human expertise during training. To further support the statement, we conduct experiments on various domains in Figure \ref{gnnclglcomp}. Based on the empirical results, it can be observed that there is a general occurrence of GNNs learning human expertise to a certain extent across different tasks. The proposed approach, treating the expertise logic as guidance, can improve GNNs to learn human expertise and further boost the model performance. The observation leads to an empirical conclusion: \textit{introducing expertise logic into graph representation learning practically improves the model performance}.

In light of such empirical finding, we propose the \textit{\underline{C}ausal \underline{L}ogic \underline{G}raph Representation \underline{L}earning}, dubbed \textit{CLGL}, to improve the model to learn the expertise logic that is causally related to the graph representation learning tasks. CLGL achieves the aforementioned objectives by enabling the GNN model to learn knowledge from higher-order causal models constructed based on expertise logic. We also conduct a theoretical analysis of CLGL based on the Structural Causal Model (SCM) \cite{DBLP:journals/ijon/Shanmugam01, pearl2009causal, glymour2016causal}, demonstrating its effectiveness. Moreover, we further incorporate the concept of interchange intervention from causal theory-related methods, enabling CLGL to facilitate a more comprehensive learning of expertise logic. The sufficient comparisons on various datasets, including the crafted and real-world datasets, support the consistent effectiveness of CLGL. 

\textbf{Contributions}:

\begin{itemize}

    \item We discover an intriguing observation, which leads to an empirical conclusion: introducing expertise logic into graph representation learning improves the model performance. Additionally, we undertake a thorough investigation, employing comprehensive theoretical analyses and rigorous proofs, to substantiate and elucidate this conclusion.
    
    % We made an interesting discovery: incorporating expertise logic into graph representation learning enhances model performance. To better understand the underlying reasons behind this observation, we build a structural causal model (SCM) and conduct comprehensive theoretical analyses and proofs.

    \item We introduce a novel CLGL method, which enables the GNN model to learn expertise logic that is causally associated with the graph representation learning task, thereby refining its performance.

    \item Extensive empirical evaluations on various datasets, including crafted and benchmark datasets, demonstrate the effectiveness of the proposed CLGL.
    
\end{itemize}

\section{Related Works}

% \subsection{Graph Neural Networks (GNNs)}
% GNNs introduce the mechanism of the neural network into graph representation learning. Since the first GNN model was proposed, multiple variants have been developed, including GCN \cite{kipf2016semi}, GAT \cite{velivckovic2017graph}, GIN \cite{xu2018powerful}, etc. These models perform graph representation learning by propagating information between adjacent nodes. Furthermore, to enhance performances and adapt to more downstream tasks, paradigms widely used by other kinds of deep learning architectures have also been applied to GNNs, such as unsupervised learning \cite{you2020does, xu2021infogcl}, semi-supervised learning \cite{baranwal2021graph}, meta learning \cite{DBLP:conf/ijcai/GaoLQS0Z22}, and reinforcement learning \cite{jiang2019graph}. 

\paragraph{Causal Learning.} Causal learning employs statistical causal inference techniques \cite{glymour2016causal} to uncover the causal relationships among observable variables, thereby enabling the discovery of more fundamental connections between data and ground-truth information. Integrating causal learning into deep learning algorithms is a growing area of interest. Researchers are exploring how causal models can enhance the performance, interpretability, and fairness of deep learning models \cite{krueger2021out,geiger2022inducing,DBLP:conf/icml/ZhouLZZ22,DBLP:conf/cvpr/LinDWZ22}. This includes developing methods to incorporate causal assumptions \cite{geiger2022inducing}, leveraging causal structures in deep learning architectures \cite{DBLP:conf/nips/0001ZDZ22}, and incorporating causal explanations into model predictions \cite{DBLP:conf/aaai/LiCM22}.

% Paradigms of causal inference such as intervention and counterfactual have been applied to various deep learning tasks \cite{krueger2021out,geiger2022inducing}. 

\paragraph{Application of Causal Inference within GNNs.} 

% As GNNs revolutionized graph representation learning by incorporating neural network mechanisms \cite{kipf2016semi, xu2018powerful, DBLP:conf/aaai/RanjanST20, miao2022interpretable}, paradigms commonly employed in other deep learning architectures have been adapted and applied to graph representation learning \cite{you2020does, xu2021infogcl, DBLP:conf/aaai/CoorayC22,baranwal2021graph, DBLP:conf/nips/YueZZL22,DBLP:conf/ijcai/GaoLQS0Z22,jiang2019graph}. However, neural network-based graph representation learning methods face challenges in explicitly discovering causal relationships. Therefore, causal inference methods have been applied to GNN-based graph representation models. For instance, in enhancing the interpretability of GNN models \cite{prado2022survey} and in improving the causality of models through data intervention \cite{DBLP:conf/iclr/WuWZ0C22}. We adopt causal learning methods to help inject expertise logic into GNNs. 

GNNs have revolutionized graph representation learning by incorporating neural network mechanisms \cite{kipf2016semi, xu2018powerful, DBLP:conf/aaai/RanjanST20, miao2022interpretable}. As a result, various paradigms commonly employed in other deep learning architectures have been adapted and applied to graph representation learning \cite{you2020does, baranwal2021graph, xu2021infogcl,  DBLP:conf/ijcai/GaoLQS0Z22, DBLP:conf/aaai/CoorayC22}. Despite their achievements, neural network-based graph representation learning approaches encounter difficulties in explicitly uncovering causal relationships. To overcome this limitation, researchers have integrated causal inference techniques into GNN-based graph representation models. These techniques have been utilized to enhance the interpretability of GNN models \cite{prado2022survey} and facilitate the incorporation of causality through data intervention \cite{DBLP:conf/iclr/WuWZ0C22}. By leveraging causal learning methods, our objective is to imbue GNNs with expertise logic.

\section{Methodology}

\begin{figure}[ht]
	\centering
    \includegraphics[width=0.8\textwidth]{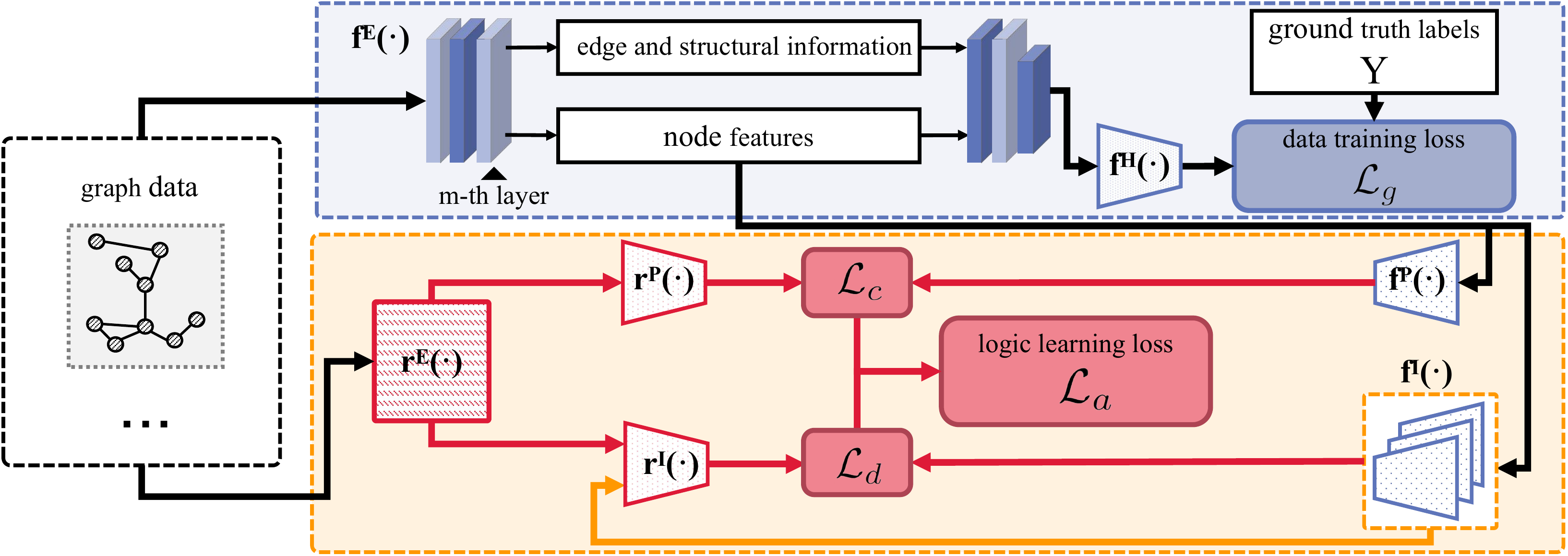} \\
	\vskip -0.05in
	\caption{The overall framework of CLGL.}
    \label{fig:framework}
	\vskip -0.2in
\end{figure}
% background knowledge, logical truth, causal reality, background facts, background causal fact, graph causal factor

\subsection{Learning Expertise Logic}

\label{sub:LK}

Before presenting our approach, we will offer a precise description of learning expertise logic. In order to achieve this, we first define two key concepts: \textit{graph causal factor} and \textit{embodiment.}

\begin{definition}
\label{df:ck}
(Graph causal factor). In graph representation learning, a graph causal factor describes the trigger for the formation of certain graph data. Since graph data is a reflection of a certain system (existing in the real world or artificially constructed), its values can necessarily be traced back to reasons found in such systems. Graph causal factors can be viewed as descriptions of such reasons. 

% In graph representation learning, a \textit{graph causal factor} refers to a fact that can cause certain graph data (which is relevant to downstream tasks) to take specific values. Furthermore, this fact cannot be directly observed from the graph data. All values in the graph data that are relevant to downstream tasks can be corresponded to their respective \textit{graph causal factors}.
\end{definition}

% As an example, in graph learning, specific graph structures can carry specific meanings and often have downstream task implications. Among these structures, cyclic structure denote a structure where nodes are connected in a circular fashion without any extra edges. Therefore, the information "a cyclic structure exists in a graph" can represent a graph causal factor. Such information cannot be directly represented as a specific numerical value within the graph data. However, it determines that certain data in the graph must conform to the cyclic structure, which means the graph causal factor "a cyclic structure exists in a graph" cause certain graph data formation.

We illustrate Definition \ref{df:ck} with a concrete example. In graph representation learning tasks, the structure of the graph is crucial for certain downstream tasks. Therefore, ``the topological structure of the graph'' can be a graph causal factor. It will cause certain graph data formation, e.g., if the value of ``the topological structure of the graph'' is ``cyclic'', then the whole graph will hold a cyclic structure. Such structure can give rise to certain properties, which can result in the assignment of particular labels to the graph, and subsequently impact downstream classification tasks.

% Let's assume cyclic structure is one of such structures. Cyclic structure denotes a structure where nodes form a closed loop without any additional edges. The existence of a cyclic structure in a graph can give rise to certain specific properties, which can result in the assignment of particular labels to the graph, and subsequently impact downstream classification tasks. Therefore, ``weather a cyclic structure exists in a graph or not'' can represent a graph causal factor, because it causes the formation of certain graph data, and cyclic structure is related to the downstream tasks. 

% Moreover, such information can not be represented as a specific numerical value within the graph data.

\begin{definition}
\label{df:e}
(Embodiment). If certain graph data is under the influence of a graph causal factor, then we call such graph data as an embodiment of the graph causal factor.
\end{definition}

For instance, in the example above, the value of ``the topological structure of the graph'' decide the structure of the whole graph. Therefore, the whole graph can be viewed as an embodiment of the graph causal factor ``the topological structure of the graph''. With the concepts graph causal factor and embodiment, we give a formal definition of \textit{expertise logic}.

\begin{definition}
\label{df:expl}
(Expertise logic). Expertise logic refers to the knowledge that can be used to discern the values of graph causal factors based on their embodiments, and this discernment can be formally represented.
\end{definition}

We carry on adopting the example above to explain expertise logic. We assume the value of ``the topological structure of the graph'' can be discerned based on its embodiments using a specific method. If this analytical method can be represented as a fixed model, denoted as $r(\cdot)$, then the knowledge used to design $r(\cdot)$ is called expertise logic, and $r(\cdot)$ can be viewed as a higher-order causal model which reflects causal relationships in the real world.

% If every expertise logic needed to formally represent the discernment of all the causal graph facts of a certain graph learning task is known, then the downstream tasks can be accomplished based on a precise mathematical formula. However, this situation is not realistic. 

However, in real-world tasks, expertise logic about most graph causal factors, including ``the topological structure of the graph'', can be difficult to acquire. In fact, only a small portion of expertise logic regarding graph causal factors can be obtained. Nevertheless, even in such cases, it can still immensely benefit graph learning. Intuitively, if a model can acquire expertise logic about certain graph causal factors $\widetilde{S}$, with a value space $\mathcal{\widetilde{S}}$, its performance and generalization on downstream tasks can be enhanced. Theoretically, we provide Theorem \ref{th:c} to show that learning information about $\widetilde{S}$ can effectively mitigate the influence of confounders.

% Our objective is to train a GNN $f(\cdot)$ that can acquire, to the greatest extent possible, information concerning the graph causal factors. Unfortunately, figure out all the graph causal factors related to downstream tasks is not feasible. However, it is possible that we hold knowledge about a small amount of graph causal factors, we denote that small amount of graph causal factors as $\widetilde{S}$. We adopt the knowledge about $\widetilde{S}$ to construct a high-level causal models $r(\cdot)$, which will aid the GNN $f(\cdot)$ in learning $\widetilde{S}$. The high-level causal models $r(\cdot)$ is able to make prediction about $\widetilde{S}$ based on the embodiment of $\widetilde{S}$. Furthermore, $r(\cdot)$ is able to tell the embodiment of $S$ from graph data $G$, i.e., the input of $r(\cdot)$ is $G$ and the output of $r(\cdot)$ is the prediction about $\widetilde{S}$.

Our goal is to train a GNN, denoted as $f(\cdot)$, that can effectively learn the expertise logic about $\widetilde{S}$ in graph data $G$ from a high-level causal model $r(\cdot)$. However, obtaining expertise logic that is directly related to downstream tasks is challenging, and such logic would be limited to functioning effectively only on specific datasets. In this paper, we pick the expertise logic that can be applied to a wide range of graph learning tasks to design $r(\cdot)$. Specifically, $r(\cdot)$ is able to make an accurate discriminative prediction about $\widetilde{S}$ based on $G$. For $r(\cdot)$, $G$ serves as the input, and the prediction of $\widetilde{S}$ is the output. We denote the output of $r(\cdot)$ as $\widetilde{T}$, $\widetilde{T} \sim p(t)$. As our emphasis lies on devising a method to learn knowledge from $r(\cdot)$, rather than determining the optimal $r(\cdot)$, we leave the implementation details of $r(\cdot)$ in Appendix \ref{apx:r}.

With $r(\cdot)$, our objective is to enable $f(\cdot)$ to possess an internal causal structure that realizes $r(\cdot)$, so as to learn the expertise logic about $\widetilde{S}$. To do so, we enforce $f(\cdot)$ to output a prediction of $\widetilde{S}$. We denote such prediction as $T$, $T \sim q(t)$. $\widetilde{T}$ and $T$ share the same value space $\mathcal{T}$. Intuitively, if given $\widetilde{S}$, the output distributions of $f(\cdot)$ and $r(\cdot)$ are the same, i.e., the following equation holds:
\begin{gather}
% \label{eq:ptc_sgsr}
p(t|\widetilde{s})=q(t|\widetilde{s}), t \in \mathcal{T}, \widetilde{s} \in \mathcal{\widetilde{S}},
\end{gather} 
then $f(\cdot)$ has learned effectively enough knowledge about $\widetilde{S}$ that $r(\cdot)$ possesses. Theorem \ref{th:ori} gives a theoretical justification for such intuition. We utilize the KL divergence to assess the distance between $p(t|\widetilde{s})$ and $q(t|\widetilde{s})$, and formulate our training objective as follows:
\begin{gather}
\mathcal{L}_{o} = KL\Big(p(t|\widetilde{s}) || q(t|\widetilde{s})\Big).
\label{eq:Lo}
\end{gather} 

However, $p(t|\widetilde{s})$ and $q(t|\widetilde{s})$ can not be directly calculated. Furthermore, $r(\cdot)$ contains only a small portion of the knowledge related to the downstream tasks, while $\widetilde{S}$ represents only a tiny subset of all graph causal factors. If the outputs of $f(\cdot)$ and $r(\cdot)$ are forced to be identical, $f(\cdot)$ would be unable to learn other crucial knowledge present in the training dataset. These issues warrant resolution.

We first address the issue that $p(t|\widetilde{s})$ and $q(t|\widetilde{s})$ can not be directly calculated. Our solution is letting the models directly predict $p(t|\widetilde{s})$ and $q(t|\widetilde{s})$. We reform $r(\cdot)$ and $f(\cdot)$ to achieve such a goal. Specifically, we reduce $r(\cdot)$ into $r^{E}(\cdot)$ that only outputs an embedding based on the input $G$. We also divided $f(\cdot)$ into $f^{E}(\cdot)$ and $f^{H}(\cdot)$. The output of $f^{E}(\cdot)$ are the representations from all GNN layers of all nodes in $G$, while $f^{H}(\cdot)$ project the last layer's representation into output prediction for downstream tasks. Then, we design two projection heads, $r^{P}(\cdot)$ and $f^{P}(\cdot)$. $r^{P}(\cdot)$ is designed based on expertise logic about $\widetilde{S}$. Given graph sample $G_{i}$, $r^{P}(\cdot)$ will predict $p_{i}(t)$  based on the embedding $r^{E}(G_{i})$. $f^{P}(\cdot)$ is a learnable neural network, which output a estimation $\hat{q}_{i}(t)$ of ${q}_{i}(t)$ based on the embedding $f^{E}(G_{i})$. $p_{i}(t)$ and $q_{i}(t)$ denote the probability density function of $\widetilde{T}$ and $T$ given $G_i$. The detailed implementation of $r^{E}(\cdot)$ and $r^{P}(\cdot)$ can be found in Appendix \ref{apx:re} and \ref{apx:rp}.

Then, we solve the problem that $f(\cdot)$ may be unable to learn other crucial knowledge present in the training dataset. We offer a solution by only utilizing the portion of the representation that is related to the embodiment of $\widetilde{S}$ to learn the corresponding expertise logic, thereby reducing the scope of influence and preserving more flexibility. Specifically, $f^{P}(\cdot)$ is designed to separate the embodiments of $\widetilde{S}$ from $G$ based on expertise logic, then input the $m$-th layer representation of the embodiment graph nodes that $f^{E}(\cdot)$ outputs, $m$ is a hyperparameter. For detailed implementation, please refer to Appendix \ref{apx:fp}. $f^{P}(\cdot)$ can be replaced with ordinary projection head $f^{H}(\cdot)$ to learn other knowledge from the dataset. 
Therefore, we can redefine our loss as $\mathcal{L}_{c}$ :
\begin{gather}
\mathcal{L}_{c} = \sum_{i}^{N} KL\Big(p_{i}(t|\widetilde{s}) || \hat{q}_{i}(t|\widetilde{s})\Big).
\label{eq:Lc}
\end{gather}

% $f^{P}(\cdot)$ can be replaced with ordinary projection head $f^{H}(\cdot)$ to learn other knowledge from the dataset. To further ensure $f^{E}(\cdot)$ can still learn other information from the training dataset, $f^{P}(\cdot)$ separate the embodiment of $\widetilde{S}$ from $G$ based on prior knowledge, and input the $m$-th layer representation of the embodiment graph nodes that $f^{E}(\cdot)$ outputs, $m$ is a hyperparameter. For detailed implementation, please refer to \ref{}. 

To conduct the calculation with our proposed model, Equation \ref{eq:Lc} can be reformulated as:
\begin{gather}
\mathcal{L}_{c} = \sum_{i=1}^{N} \mathcal{D}_{KL}\Big(r^{P}\big(r^{E}(G_i)\big), f^{P}\big(f^{E}(G_i)\big)\Big),
\label{eq:Lc2}
\end{gather} 
where $\mathcal{D}_{KL}(\cdot)$ calculates the KL divergence of two input distributions via a discrete manner. $N$ is the number of graph samples. The validity of Equation \ref{eq:Lc2} was proved with Corollary \ref{corollary:Lc2}. 

% Similarly,  define the outputs of $r(\cdot)$ and $f(\cdot)$ as embeddings, excluding the projection heads that project the embeddings into the final output. Then, we specifically design two projection heads, $p^{R}(\cdot)$ and $p^{F}(\cdot)$. $p^{R}(\cdot)$ will project $r(G)$ into a discrete probability distribution of possible values for the prediction of $\widetilde{S}$. 

%is the graph data, and the output representation of  which will predict the probability of each value of t 

\subsection{Enhancement with Interchange Interventions}

The design outlined above presents a feasible approach for enabling the GNN model to learn the expertise logic concerning $\widetilde{S}$, in which only the embodiments of $\widetilde{S}$ are involved in the learning process. However, in the actual training dataset, these embodiments may not exhaust all possible values of $\widetilde{S}$ or offer rather a small amount of training cases for certain values, which will lead to the knowledge contained in the high-level causal model can not be fully studied by the GNN model. To ensure that the GNN model can fully acquire such knowledge, we employ \textit{interchange intervention} \cite{geiger2022inducing, DBLP:conf/emnlp/GeigerCKP19}, a method based on causal theory, to facilitate and improve the learning process.

\paragraph{Interchange Intervention.} For a model $\mathcal{M}$ that is used to process two different input samples, $X_{a}$ and $X_{b}$, the interchange intervention conducted on a model $\mathcal{M}$ can be viewed as providing the output of the model $\mathcal{M}$ for the input $X_{a}$, except the variables $V$ are set to the values they would have if $X_{b}$ were the input. In our task, we adopt the node representations outside of the embodiments of $\widetilde{S}$  to conduct interchange intervention, as they do not participate in expertise logic learning before.
% With the aforementioned design, we hold a proper way to learn the prior knowledge about $\widetilde{S}$. However, given that prior knowledge about $\widetilde{S}$ is challenging to obtain, in practical tasks, the set of values for $\widetilde{S}$ that can be thoroughly analyzed is often quite small, perhaps only including one or two distinct values. In such circumstances, employing the method designed earlier directly will lead to a high risk of trivial solutions arising. To address such an issue and to ensure the knowledge contained in $r(\cdot)$ can be learned fully, we employ a causality-based method called \textit{interchange intervention} \cite{Inducing Causal Structure for Interpretable Neural Networks, Posing fair generalization tasks for natural language inference} to help enhance the learning process.

We conduct interchange intervention on the GNN model by modifying the projection head $f^{P}(\cdot)$ into $f^{I}(\cdot)$. Different from $f^{P}(\cdot)$, which inputs only the node representations within the embodiment of $\widetilde{S}$, $f^{I}(\cdot)$ will pick a certain amount of node representations outside of the embodiment of $\widetilde{S}$ and replace the original ones. 

Such interchange intervention, similar to conventional intervention, can be viewed as a modification to the whole system \cite{pearl2009causality}. Our desideratum is that the output of the GNN and the high-level causal models are the same under identical interchange interventions. However, since these two models adopt different structures, identical interchange interventions are hard to conduct. Therefore, we follow \cite{pearl2011graphical} and treat the interchange intervention on the GNN model as a variable. Thus, we can adjust the output of the high-level causal model based on such a variable accordingly. Since the higher-order causal model is built on pre-acquired expertise logic, we can adopt the same logic to predict what output it should produce given specific interchange intervention. Formally, we modify the projection head $r^{P}(\cdot)$ to $r^{I}(\cdot)$, which can be defined as follows:

\begin{gather}
r^{I}\Big(r^{E}(G), f^{I}(\cdot)\Big) = Normalize\Big(r^{P}\big(r^{E}(G)\big) \circ h\big(f^{I}(\cdot)\big)\Big),
\label{eq:ri}
\end{gather} 

where $h(\cdot)$ judges the type of $f^{I}(\cdot)$, and will output weights based on such judgement. $\circ$ denotes the Hadamard Product. $Normalize(\cdot)$  represents the normalization operation. During learning, we can modify $f^{I}(\cdot)$ to conduct different interchange interventions. The training objective $\mathcal{L}_{d}$ under interchange intervention can be represented as:

\begin{gather}
\mathcal{L}_{d} = \sum_{i=1}^{N}{ \sum_{j=1}^{K}{ \mathcal{D}_{KL}\Big(r^{I}\big(r^{E}(G_{i}), f_{j}^{I}(\cdot)\big), f^{I}_{j}\big(f^{E}(G_i)\big)\Big)}},
\label{eq:Li}
\end{gather} 

where $K$ denote the total number of the types of $f^{I}(\cdot)$, $f_{j}^{I}(\cdot)$ denote the $j$-th type. The detailed implementation of  $r^{I}(\cdot)$ can be found in Appendix \ref{apx:ri}.

\subsection{Training Procedure}
The overall framework is illustrated in Figure \ref{fig:framework}. For learning expertise logic, we define our training objective with both $\mathcal{L}_{c}$ and $\mathcal{L}_{d}$:
\begin{gather}
\mathcal{L}_{a} = \mathcal{L}_{c} + \lambda \mathcal{L}_{d},
\label{eq:L2}
\end{gather}
$\mathcal{L}_{a}$ denotes the total loss for expertise logic learning, $\lambda$ is a hyperparameter that balances the influence of learning under interchange intervention. Besides expertise logic, our GNN model is also trained with conventional labeled data. The training object of which can be formulated as:
\begin{gather}
	 \mathcal{L}_{g} = \mathcal{H}\Big(  f^{H}\big(f^{E}(G)\big), Y\Big),
	\label{eq:lg0}
\end{gather}
where $\mathcal{H}(\cdot)$ calculates the cross entropy loss, and $f^{H}(\cdot)$ denotes the projection head for label prediction. We update the model alternatively with $\mathcal{L}_{g}$ and $\mathcal{L}_{a}$. Only $f^{H}(\cdot)$ and $f^{E}(\cdot)$ are utilized for performance test.

\section{Theoretical Analysis with Causality and Information Theory}
\label{s:tac}

% In this section, we conduct theoretical analysis on our method based on causality and information theory to provide proofs for its validity.

% \subsection{SCM} 

% SCM \cite{DBLP:journals/ijon/Shanmugam01} is a framework used to describe how nature assigns values to variables of interest. Formally, SCM consists of two sets of variables, $U$ and $V$, and a function set $F$ that define the relations between different variables in the SCM. The variables in $U$ are called exogenous variables. These variables will not be affected by the internal variables $V$ of the SCM. Therefore, when using SCM for analysis, the causes of variables in $U$ will not be discussed. Each SCM can be represented as a graphical model that consists of a set of nodes representing the variables within $U$ and $V$, and a set of edges between the nodes representing the functions within $F$. We employ SCM to analyze the graph representation learning from a causal perspective.

% \begin{figure}[ht]
% 	\centering

% 	\subfigure[]{
% 		\begin{minipage}{0.22\textwidth} 
% 			\includegraphics[width=\textwidth]{SCM.pdf} \\
%             \label{{fig:SCMa}
% 		\end{minipage}
%     }\hspace{2mm}
% 	\subfigure[]{
% 		\begin{minipage}{0.22\textwidth} 
% 			\includegraphics[width=\textwidth]{SCMb.pdf} \\
%             \label{{fig:SCMb}
% 		\end{minipage}
% 	}

%      \caption{Graphical model of the proposed SCM.}
%      \label{{fig:SCM}
 
% 	%\vskip -0.1in
% \end{figure}

\subsection{Causal modeling} 
\label{subsec:cm}

% \begin{figure}[ht]
% 	\centering
% 	\subfigure[]{
% 		\begin{minipage}{0.18\textwidth} 
% 			\includegraphics[width=\textwidth]{IC-d.pdf} \\
%             \label{fig:SCM}
% 		\end{minipage}
%     }\hspace{2mm}
% 	\subfigure[]{
% 		\begin{minipage}{0.18\textwidth} 
% 			\includegraphics[width=\textwidth]{SCM-b.pdf} \\
%             \label{fig:SCMb}
% 		\end{minipage}
% 	}

% 	% \vskip -0.05in
% 	\caption{Graphical representations of the SCMs. Subsection \ref{subsec:cm} gives detailed explanations of these SCMs. }
% 	\label{fig:IC}
% 	%\vskip -0.1in
% \end{figure}

\begin{wrapfigure}{r}{0.2\textwidth}
    \vskip -0.15in
    \centering
    \includegraphics[width=0.2\textwidth]{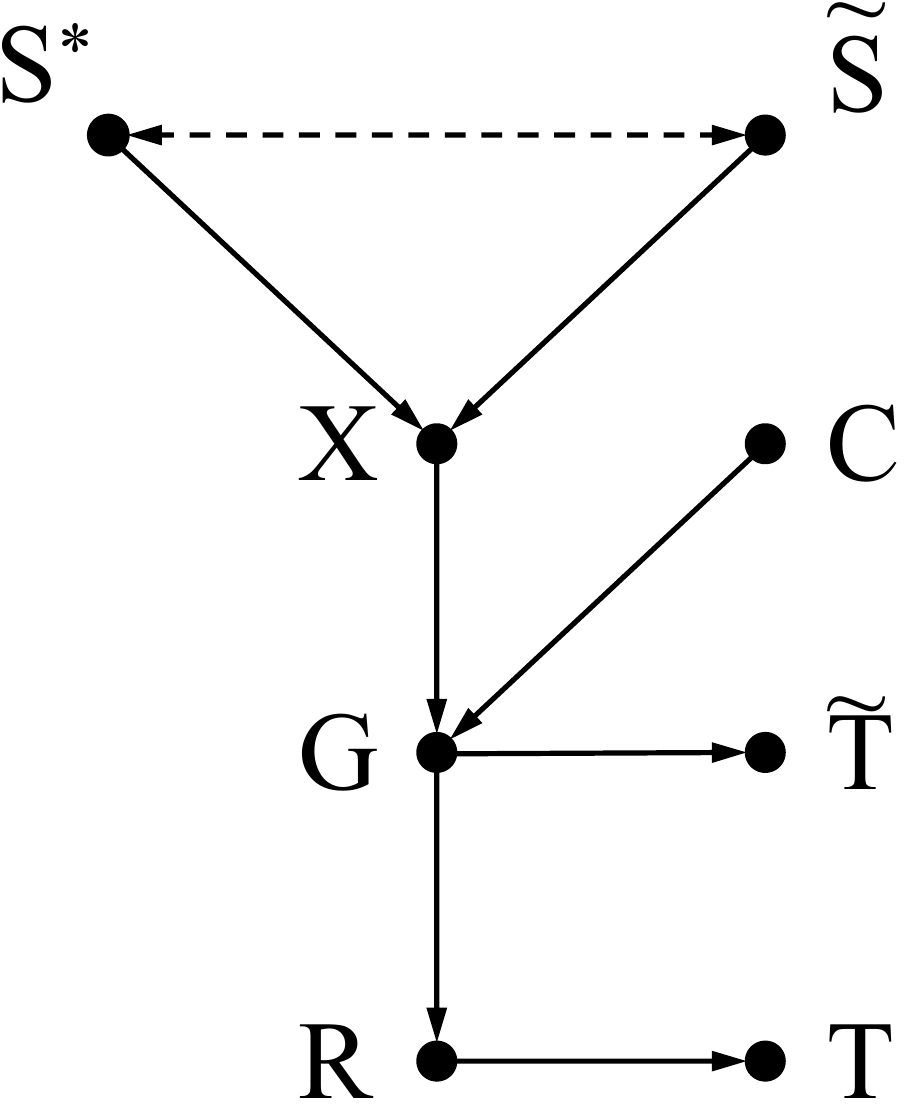}
    \vskip -0.05in
    \caption{Graphical representation of the SCM.}
    \vskip -0.2in
    \label{fig:SCM}
\end{wrapfigure}

\paragraph{SCM.} We adopt SCM to conduct causal modeling. SCM \cite{DBLP:journals/ijon/Shanmugam01} is a framework utilized to describe the manner in which nature assigns values to variables of interest. Formally, an SCM consists of variables and the relationships between them. Each SCM can be represented as a graphical model consisting of a set of nodes representing the variables, and a set of edges between the nodes representing the relationships. 

% We use SCM to conduct causal modeling. SCM \cite{DBLP:journals/ijon/Shanmugam01} is a framework utilized to describe the manner in which nature assigns values to variables of interest. Formally, an SCM comprises two sets of variables, $U$ and $V$, and a set of functions $F$ that define the relationships between different variables within the SCM. The variables in $U$ are referred to as exogenous variables, as they are not affected by the internal variables $V$ of the SCM. Each SCM can be represented as a graphical model consisting of a set of nodes representing the variables within $U$ and $V$, and a set of edges between the nodes representing the functions within $F$. Therefore, when utilizing SCM for analysis, the causes of variables in $U$ are not taken into consideration.

We formalize the candidate problem by using an SCM illustrated in Figure \ref{fig:SCM}. The validity of this SCM is proved based on the Inductive Causation (IC) algorithm \cite{verma1990equivalence}, as demonstrated in Appendix \ref{apx:SCM}. Among the SCM, the variables $S^{*}$ and $\widetilde{S}$ represent the graph causal factors, which hold a direct causal relationship with the graph data. $\widetilde{S}$ denotes the accessible graph causal factors that can be discerned with the available expertise logic. $S^{*}$ denotes the rest graph causal factors. The variable $X$ represents the embodiments of $S^{*}$ and $\widetilde{S}$ within the graph data. For ease of understanding such concepts, we establish an intuitive illustration: $\widetilde{S}$ can present the topology structure of a graph, $S^{*}$ are the other properties of the graph. If the value of $\widetilde{S}$ is "cyclic", then its embodiment, which is contained in $X$, will take a cyclic structure. Please note that this is merely a simplified example for illustrative purposes, and obtaining all the relevant expertise logic for the whole structure of a graph in practical tasks is not feasible. $C$ represents any confounding factors present within the data, $G$ represents the graph data itself, and $R$ represents the learned representation of the graph data. $\widetilde{T}$ and $T$ are the outputs of the high-level causal model and the GNN. The links in Figure \ref{fig:SCM} are as follows: 

\begin{itemize}
\item $S^{*} \to X, \widetilde{S} \to X.$ $X$ denotes the embodiment of $S^{*}$ and  $\widetilde{S}$ within the graph data.
\item $X \to G \gets C.$ The graph data $G$ consists of two parts: X and confounder $C$.
\item $G \to R.$ A GNN encoder encodes $G$ into representation $R$ that consists of the node representations of different layers and the graph representation.
\item $S^{*} \dashedleftarrow \dashedrightarrow \widetilde{S}.$ The bidirectional arrow with a dashed line indicates that the causal relationship between the two cannot be confirmed.
\item $G \to \widetilde{T} , R \to T.$ $\widetilde{T}$ and $T$ are variables that calculated based on $G$ and $R$.
\end{itemize}

\subsection{Analysis based on the causal models} 

Based on SCM in Figure \ref{fig:SCM}, we provide theoretical proof to support the intuition that acquiring prior knowledge about graph causal factors is beneficial for enhancing a model's performance. To achieve such a goal, We propose the following theorem.

\begin{theorem}
\label{th:c}
For a graph representation learning process with a causal structure represented by the SCM in Figure \ref{fig:SCM}, increasing the mutual information $I(R;\widetilde{S})$ between $R$ and $\widetilde{S}$ can decrease the upper bound of the mutual information $I(R;C)$ between $R$ and $C$. Formally:
\begin{gather}
I(R;C) \leq 1-I(R;\widetilde{S}).
\end{gather} 
\end{theorem}

The proof of Theorem \ref{th:c} can be found in Appendix \ref{apx:pfthc}. Theorem \ref{th:c} establishes a direct relationship between a model's understanding of graph causal factors and the reduction of confounding influence. As the model gains a deeper understanding of graph causal factors, the confounding influence diminishes. Next, we prove the validity of the training objective proposed by Equation \ref{eq:Lo}.

\begin{theorem}
\label{th:ori}
For a certain graph learning process with a causal structure represented by the SCM in Figure \ref{fig:SCM}, $\widetilde{T} \sim p(t), T \sim q(t), t\in\mathcal{T}$, if the high-level causal models $r(\cdot)$ is effective enough such that $I(\widetilde{T}, \widetilde{S}) = I(G, \widetilde{S})$, then $I(R, \widetilde{S})$ is maximized if $p(t|\widetilde{s}) = q(t|\widetilde{s})$.
\end{theorem}

The proof of Theorem \ref{th:ori} can be found in Appendix \ref{apx:pfori}. According to Theorem \ref{th:ori}, if Equation \ref{eq:Lo} holds, $I(R, \widetilde{S})$ will be maximized, which indicates that the model has learned the maximum amount of knowledge about $\widetilde{S}$. With Theorem \ref{th:ori}, we present a corollary to substantiate the validity of $\mathcal{L}_{c}$.

\begin{corollary}
\label{corollary:Lc2}
For a specific graph dataset $G$, if the training samples in $G$ sufficiently cover all embodiments corresponding to $\widetilde{S}$, and the $\mathcal{L}_{c}$ which is defined by Equation \ref{eq:Lc2} reaches zero, then the maximized value of $I(R, \widetilde{S})$ stated in Theorem \ref{th:ori} can be achieved.
\end{corollary}

The proof is presented in Appendix \ref{apx:pfc}.

% \begin{corollary}
% \label{crl:lc}
% Given the SCM and the assumptions outlined in Theorem \ref{th:ori}, given large enough training data, then the following equation holds:
% \begin{gather}
% KL\Big(p(t|\widetilde{s}) || q(t|\widetilde{s})\Big) = 0 ,
% \label{eq:Lo}
% \end{gather} 
% if and only if the following equation holds:
% \begin{gather}
% \sum_{i=1}^{N} \mathcal{D}_{KL}\Big(r^{P}(r^{E}(G_i)), f^{P}(f^{E}(G_i))\Big) = 0.
% \label{eq:Lc}
% \end{gather} 
% \end{corollary}

\section{Experiments}

In this section, we carry out multiple experiments aimed at addressing the research questions:

\textbullet \ \textbf{RQ1:} How effective is CLGL in learning expertise logic? Furthermore, Can such expertise logic be applied to various datasets?

\textbullet \ \textbf{RQ2:} How does CLGL differ from conventional methods? How does the knowledge of expertise logic specifically assist in such a training process?

% \item We introduce a novel CLGL method, which enhances the model's ability to learn expertise logic that is causally associated with the graph representation learning task, thereby refining its performance.

% \item Extensive empirical evaluations on various datasets, including crafted and benchmark datasets, demonstrate the effectiveness of the proposed CLGL.

\subsection{Settings}
\paragraph{Datasets.}
For experiments, we employed two distinct types of datasets, one comprising comprehensive topological structure information and the other containing rich node attribution information. 

For the topological structure information abundant datasets, we developed a high-level causal model based on general topology prior knowledge and trained it on multiple datasets, including:1) Spurious-Motif, a synthetic Out Of Distribution (OOD) dataset \cite{DBLP:conf/iclr/WuWZ0C22} based on the data generation method of \cite{DBLP:conf/nips/YingBYZL19}; 2) Motif-Variant, also a synthetic OOD dataset that is created by ourselves using the data generation method following \cite{DBLP:conf/nips/YingBYZL19}, while the internal topological structure is different from the previous dataset. 3) The In-Distribution (ID) version of these datasets. To guarantee the fairness and efficacy of our experiments, the high-level causal model we devised did not integrate any information directly related to downstream tasks, but rather comprised only generic topological structure knowledge. Additionally, we confirmed our method's universality through validation on multiple datasets with the same high-level causal model. 

\begin{table*}[ht]\scriptsize
    \vskip -0.1in
	\setlength{\tabcolsep}{5pt}
 	\caption{Performance of classification accuracy in Spurious-Motif. Spurious-Motif (ID) denotes the ID version dataset. Bias represents the degree of distribution shift between the training set and the test set. Some of the results are cited from \cite{DBLP:conf/iclr/WuWZ0C22}. The best records are highlighted in bold.}
	\centering
		\begin{tabular}{l|cccc|c}
			\hline\rule{0pt}{6pt}

			\multirow{2}*{Method}   & \multicolumn{4}{c}{Spurious-Motif } \vline&  Spurious-Motif   \\
			\cline{2-5}\rule{0pt}{6pt} 
			& Balanced & bias = 0.5 & bias = 0.7 & bias = 0.9 & (ID) \\ 	
			%\hline
			\hline\rule{-2pt}{6pt}
			\text{ERM}  & 42.99$\pm$1.93 & 39.69$\pm$1.73 & 38.93$\pm$1.74 & 33.61$\pm$1.02 & {92.91$\pm$1.98}\\
			% \hline\rule{-3pt}{10pt}
			% \hline
			% \hline\rule{0pt}{10pt}
			\text{GAT \cite{velivckovic2017graph}} & 43.07$\pm$2.55 & 39.42$\pm$1.50 & 37.41$\pm$0.86 & 33.46$\pm$0.43 & {91.31$\pm$2.28}\\
			\text{Top-k Pool \cite{gao2019graph}}  &  43.43$\pm$8.79 & 41.21$\pm$7.05 & 40.27$\pm$7.12 & 33.60$\pm$0.91 & {90.83$\pm$3.04}\\
            \text{GSN \cite{DBLP:journals/pami/BouritsasFZB23}} &   43.18$\pm$5.65 & 34.67$\pm$1.21 & 34.03$\pm$1.69 & 32.60$\pm$1.75 & {92.81$\pm$1.10}\\
			\text{Group DRO \cite{sagawa2019distributionally}} &   41.51$\pm$1.11 & 39.38$\pm$0.93 & 39.32$\pm$2.23 & 33.90$\pm$0.52 & {92.18$\pm$1.12}\\
			\text{IRM \cite{arjovsky2019invariant}}   & 42.26$\pm$2.69 &  41.30$\pm$1.28 &  40.16$\pm$1.74 &  35.12$\pm$2.71 & {91.12$\pm$1.56}\\
			\text{V-REx \cite{krueger2021out}}  & 42.83$\pm$1.59 &  39.43$\pm$2.69 &  39.08$\pm$1.56 &34.81$\pm$2.04 & {91.08$\pm$1.85}\\
			\text{DIR \cite{DBLP:conf/iclr/WuWZ0C22}}    & 42.53$\pm$3.38 &  41.45$\pm$2.12 &  41.03$\pm$1.53 &  39.20$\pm$1.94 & {93.02$\pm$1.89}\\
			\hline\rule{-1pt}{6pt}
               CLGL-D  & 42.83$\pm$2.33 &  40.13$\pm$1.94 &  39.08$\pm$1.68 &  34.11$\pm$2.02  & 91.69$\pm$1.30 \\
   
               CLGL-A & 47.08$\pm$1.56 &  43.03$\pm$1.90 &  41.15$\pm$1.54 &  40.13$\pm$1.89   & 94.03$\pm$1.28 \\
			\textbf{CLGL}   & \bf{52.18$\pm$2.35} &   \bf{50.11$\pm$1.92} &  \bf{49.02$\pm$1.79} &  \bf{43.31$\pm$2.07}  & \bf{96.53$\pm$1.28}\\
			\hline
		\end{tabular}

	\label{tab:sm}
    \vskip -0.1in
\end{table*}

\begin{table*}[ht]\scriptsize
    \vskip -0.1in
	\setlength{\tabcolsep}{5pt}
	\centering
 	\caption{Performance of classification accuracy in Motif-Variant. Motif-Variant (ID) denotes the ID version dataset. Bias represents the degree of distribution change between the training set and the test set. The best records are highlighted in bold.}
		\begin{tabular}{l|cccc|c}
			\hline\rule{0pt}{6pt}

			\multirow{2}*{Method}   & \multicolumn{4}{c}{Motif-Variant }   \vline & Motif-Variant\\
			\cline{2-5}\rule{0pt}{6pt} 
			& Balanced & bias = 0.5 & bias = 0.7 & bias = 0.9 & (ID)\\ 	
			%\hline
			\hline\rule{-2pt}{6pt}
			\text{ERM}  & 48.18$\pm$3.46 & 46.38$\pm$2.52 & 45.78$\pm$2.81 & 42.31$\pm$2.13 & {94.78$\pm$1.07}\\
			% \hline\rule{-3pt}{10pt}
			% \hline
			%\hline\rule{0pt}{10pt}
			\text{GAT \cite{velivckovic2017graph}}  & 49.13$\pm$2.96 & 47.78$\pm$2.12 & 45.63$\pm$1.93 & 41.48$\pm$0.85 & {93.32$\pm$2.60}\\
			\text{Top-k Pool \cite{gao2019graph}} &  48.56$\pm$7.10 & 46.08$\pm$7.93 & 44.37$\pm$8.07 & 42.10$\pm$6.13 & {93.17$\pm$3.21} \\
            \text{GSN \cite{DBLP:journals/pami/BouritsasFZB23}} &   47.05$\pm$6.03 & 44.08$\pm$2.32 & 43.15$\pm$1.68 & 40.19$\pm$1.93 & {93.90$\pm$2.11}\\
			\text{Group DRO \cite{sagawa2019distributionally}}  &   44.03$\pm$1.38 & 42.06$\pm$1.21 & 41.15$\pm$1.56 & 39.90$\pm$1.02 & {92.54$\pm$1.31}\\
			\text{IRM \cite{arjovsky2019invariant}}   & 48.13$\pm$2.86 &  44.30$\pm$1.49 &  42.01$\pm$2.03 &  40.18$\pm$2.33 & {93.17$\pm$1.21}\\
			\text{V-REx \cite{krueger2021out}}  & 49.76$\pm$1.68 &  46.83$\pm$2.36 &  43.12$\pm$1.50 &  42.37$\pm$1.99  & {93.13$\pm$1.25}\\
			\text{DIR \cite{DBLP:conf/iclr/WuWZ0C22}}   & 49.66$\pm$2.85 &  47.76$\pm$2.73 &  44.80$\pm$1.32 &  42.90$\pm$1.68 & {94.46$\pm$1.57} \\
			 \hline\rule{-1pt}{6pt}
            CLGL-D  & 48.01$\pm$1.87 &  47.05$\pm$1.62 &  45.37$\pm$1.49 &  42.11$\pm$1.83  & 92.01$\pm$0.86 \\
    
    			CLGL-A  & 48.96$\pm$1.73 &  48.60$\pm$1.43 &  46.23$\pm$1.43 &  45.80$\pm$1.87  & 93.56$\pm$0.99 \\
			\textbf{CLGL}  & \bf{55.36$\pm$2.38} &  \bf{53.00$\pm$2.10} &  \bf{49.84$\pm$2.32} &  \bf{47.96$\pm$1.87} & \bf{97.57$\pm$0.92} \\
			\hline
		\end{tabular}

	\label{tab:smv}
    \vskip -0.1in
\end{table*}

\begin{table*}[ht]\scriptsize
    \vskip -0.1in
    \setlength{\tabcolsep}{5pt}
	\centering
 	\caption{Performance of classification accuracy in Graph-SST5 and Graph-Twitter. Graph-SST5 (OOD) and Graph-Twitter (OOD) denote the OOD version datasets. The best records are highlighted in bold.}
		\begin{tabular}{l|cc||cc}
			\hline\rule{0pt}{6pt}
			\multirow{2}*{Method} & \multirow{2}*{Graph-SST5} & Graph-SST5 & \multirow{2}*{Graph-Twitter} & Graph-Twitter \\ 	
			% \cline{2-5}\rule{0pt}{10pt}
			&  & (OOD) &  & (OOD) \\ 
			%\hline
			\hline\rule{-3pt}{6pt}
			\text{ERM} & 49.02$\pm$0.82 & 39.39$\pm$0.66 & 65.46$\pm$1.01 & 63.04$\pm$1.46 \\
			% \hline\rule{-3pt}{10pt}
			% \hline
			%\hline\rule{0pt}{10pt}
			\text{GAT \cite{velivckovic2017graph}} & 48.51$\pm$1.86 & 37.58$\pm$1.67 & 63.57$\pm$0.95 & 62.38$\pm$0.98 \\
			\text{Top-k Pool \cite{gao2019graph}} &  49.72$\pm$1.42 & 36.26$\pm$1.86 & 63.41$\pm$1.95 & 62.95$\pm$1.09 \\
            \text{GSN \cite{DBLP:journals/pami/BouritsasFZB23}} &  48.64$\pm$1.60 & 38.78$\pm$1.84 & 63.18$\pm$1.89 & 63.07$\pm$1.18 \\
			\text{Group DRO \cite{sagawa2019distributionally}} &   47.44$\pm$1.12 & 37.78$\pm$1.12 & 62.23$\pm$1.43 & 61.90$\pm$1.03 \\
			\text{IRM \cite{arjovsky2019invariant}}  & 48.08$\pm$1.30 &  38.68$\pm$1.62 &  64.11$\pm$1.58 &  62.27$\pm$1.55 \\
			\text{V-REx \cite{krueger2021out}} & 48.55$\pm$1.10&  37.10$\pm$1.18 &  64.84$\pm$1.46 &  63.42$\pm$1.06 \\
			\text{DIR \cite{DBLP:conf/iclr/WuWZ0C22}}   & 49.16$\pm$1.31 &  38.67$\pm$1.38 &  65.14$\pm$1.37 &  63.49$\pm$1.36 \\
			\hline\rule{-1pt}{6pt}
              CLGL-D  & 48.93$\pm$0.76 &  38.85$\pm$1.53 &  64.79$\pm$1.05 &  
               62.18$\pm$0.83  \\
   
               CLGL-A  & 49.63$\pm$0.89 &  39.00$\pm$1.25 &  65.32$\pm$0.93 &  
               61.98$\pm$0.96  \\
			\textbf{CLGL}  & \bf{50.38$\pm$0.87} &  \bf{42.13$\pm$1.50} &  \bf{66.68$\pm$0.81} &  \bf{63.98$\pm$1.20}  \\
			\hline
		\end{tabular}

	\label{tab:graph}
    \vskip -0.1in
\end{table*}

For the node attribution information abundant datasets, we developed another high-level causal model. We also ensure that the model contains only generic knowledge that does not directly relate to downstream tasks, and conduct experiments with the same high-level causal model on multiple real-world datasets, including Graph-SST5 \cite{yuan2020explainability}, Graph-Twitter \cite{yuan2020explainability}, and their OOD versions. The details of the high-level causal model and the datasets can be found in Appendix \ref{apx:r} and \ref{apx:dataset}.

\paragraph{Baselines.}
We compare our method with Empirical Risk Minimization (ERM) and various causality-enhanced methods that can be divided into two groups: 1) the interpretable baselines, including GAT ,Top-k Pool and GSN, 2) the robust learning baselines, including Group DRO, IRM, V-REx and DIR. For a fair comparison, we follow the experimental principles of \cite{DBLP:conf/iclr/WuWZ0C22} and adopt the same training setting for all models, which is described in Appendix \ref{app:set} in detail. For each task, we report the mean performance ± standard deviation over five runs. For further demonstration of the universality of our proposed method, we also construct the CLGL-D model. CLGL-D directly employs the output of the higher-order causal model $r(\cdot)$ as labels for training, along with the existing labels within the dataset. Consequently, we can assess whether the output of $r(\cdot)$ contains information directly relevant to downstream tasks that may aid the training procedure. Furthermore, to showcase the benefits of our proposed enhancement with interchange interventions, we develop the CLGL-A model to conduct corresponding ablation studies, which removes the enhancement with interchange interventions.

% To demonstrate the universality of our proposed method, we also construct the model CLGL-D. CLGL-D directly utilizes the output of our higher-order causal system $r(\cdot)$, as labels for training. Therefore, we can observe whether the output of $r(\cdot)$ contains information directly relevant to downstream tasks. To demonstrate the utility of our proposed enhancement with interchange interventions, we construct the model CLGL-A to perform corresponding ablation studies, which removes the enhancement with interchange interventions from CLGL model. 

\begin{figure*}[ht]
	\centering
	\subfigure[Accuracy and loss obtained over 200 epochs on multiple Motif-Variant datasets with distinct biases.]{
		\begin{minipage}{0.63\textwidth} 
			\includegraphics[width=\textwidth]{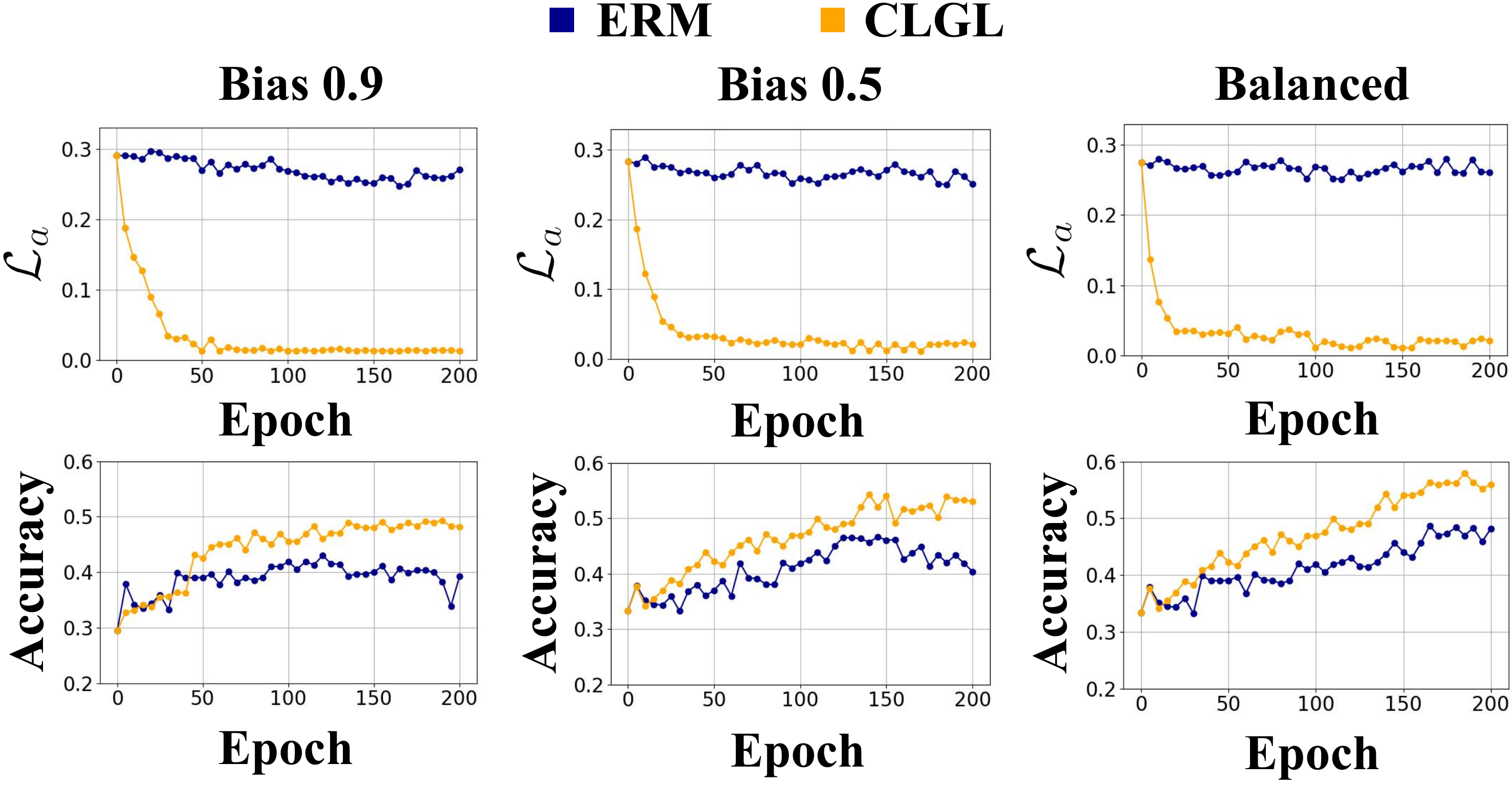} \\
%  		\subcaption{\hspace*{2.3cm}(a) MUTAG}
        \label{fig:ida}
		\end{minipage}
	}\hspace{1mm} 
 	\subfigure[Visualized features]{
		\begin{minipage}{0.33\textwidth} 
			\includegraphics[width=\textwidth]{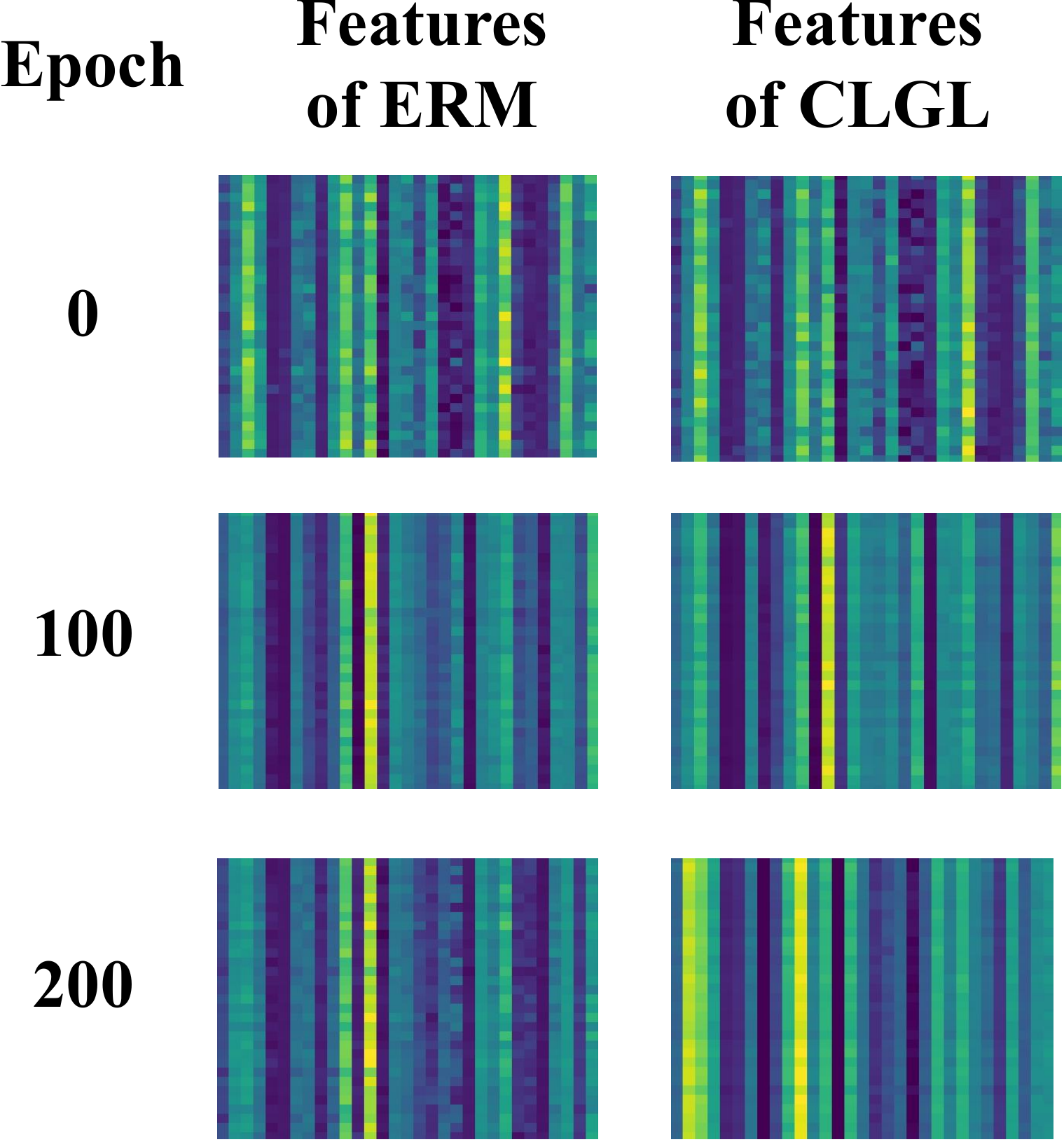} \\
%  		\subcaption{\hspace*{2.3cm}(a) MUTAG}
        \label{fig:idb}
		\end{minipage}
	}\hspace{0mm} 
 	\subfigure[Features after dimensionality reduction using t-SNE \cite{van2008visualizing}]{
		\begin{minipage}{0.33\textwidth} 
			\includegraphics[width=\textwidth]{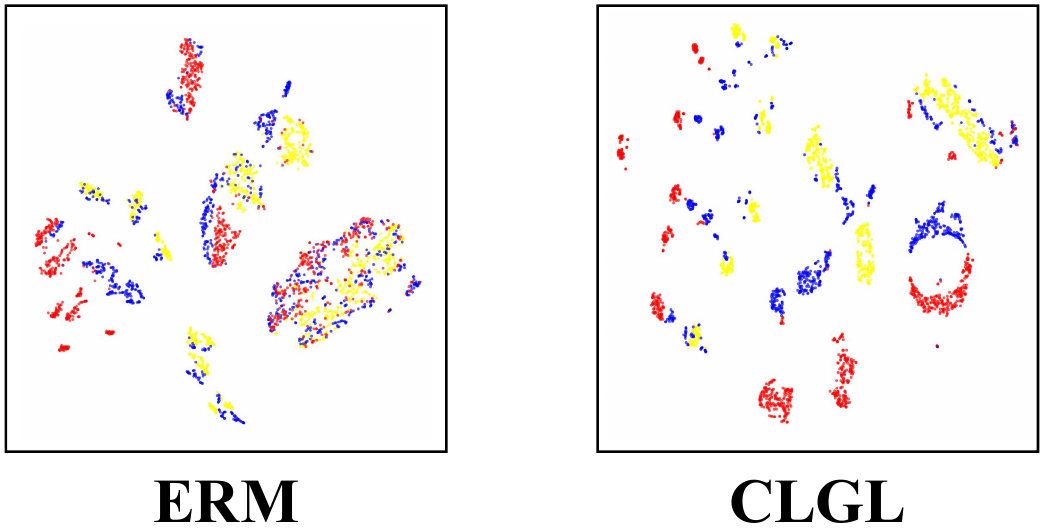} \\
%  		\subcaption{\hspace*{2.3cm}(a) MUTAG}
        \vskip -0.1in
        \label{fig:idc}
		\end{minipage}
	}\hspace{3mm} 
	\subfigure[Parameter experiments conducted on different datasets]{
		\begin{minipage}{0.60\textwidth} 
			\includegraphics[width=\textwidth]{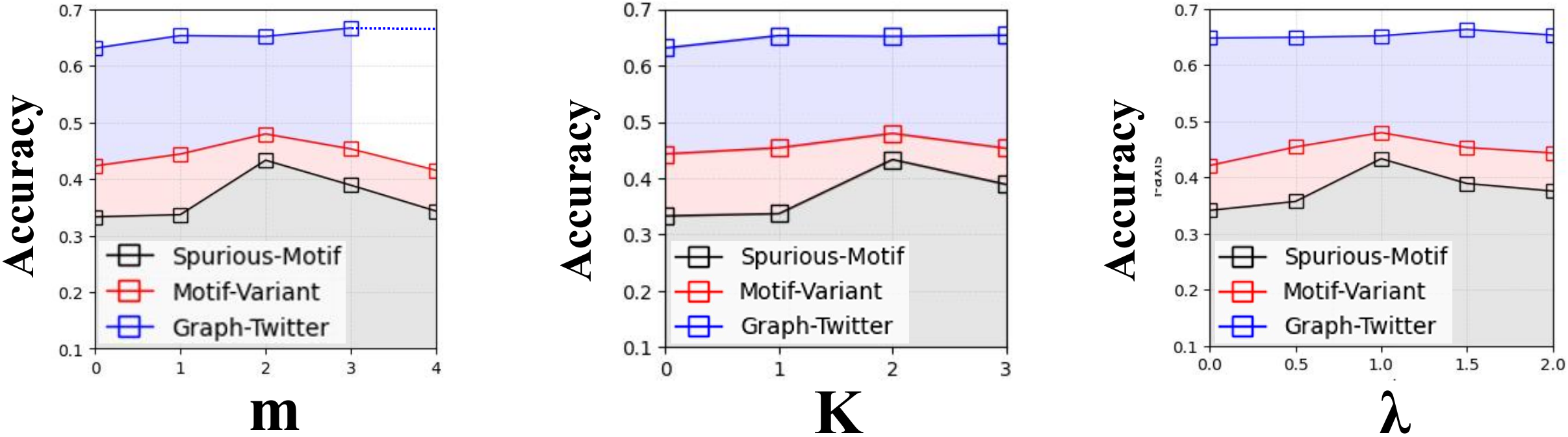} \\
%  		\subcaption{\hspace*{2.3cm}(a) MUTAG}
        \vskip -0.1in
        \label{fig:idd}
		\end{minipage}
	}\hspace{0mm} 

% 	\subfigure[IMDB-B]{
% 		\begin{minipage}{0.3\textwidth}%[b]%{0.2\textwidth}
% 			\includegraphics[width=\textwidth]{blt-c.pdf} \\
% % 		\subcaption{\hspace*{2.2cm}(c) IMDB-B}
% 		\end{minipage}
% 	}
	\vskip -0.05in
	\caption{Results of further experiments.}
	\label{fig:visual}
	\vskip -0.15in
\end{figure*}

\subsection{Comparison with State-of-the-art Methods (RQ1)}
The results are reported in Table \ref{tab:sm}, \ref{tab:smv}, and \ref{tab:graph}. We observe from the tables and find that our method outperforms all baselines on all downstream tasks by large margins. Another observable phenomenon is that CLGL significantly outperforms CLGL-D. Since the expertise logic we introduced represents \textit{universal} graph knowledge not specific to any downstream task, directly utilizing this universal knowledge cannot enhance the model's performance on downstream tasks. However, CLGL surpasses state-of-the-art performance, indicating that using graph neural networks to learn universal expertise logic can help them acquire logical inductive capabilities, enabling them to better extract discriminative information from graph data. Furthermore, it is observable that CLGL outperforms CLGL-A, which verifies the necessity of our proposed enhancement with interchange interventions.

% \begin{figure*}[ht]
% 	\centering
% 	\subfigure[]{
% 		\begin{minipage}{0.30\textwidth} 
% 			\includegraphics[width=\textwidth]{hyp-a.pdf} \\
% %  		\subcaption{\hspace*{2.3cm}(a) MUTAG}
%         \vskip -0.1in
%         \label{fig:visual-a}
% 		\end{minipage}
% 	}\hspace{3mm} 
% 	\subfigure[]{
% 		\begin{minipage}{0.55\textwidth} 
% 			\includegraphics[width=\textwidth]{hyp-b.pdf} \\
% %  		\subcaption{\hspace*{2.3cm}(a) MUTAG}
%         \vskip -0.1in
%         \label{fig:visual-b}
% 		\end{minipage}
% 	}\hspace{0mm} 
% % 	\subfigure[IMDB-B]{
% % 		\begin{minipage}{0.3\textwidth}%[b]%{0.2\textwidth}
% % 			\includegraphics[width=\textwidth]{blt-c.pdf} \\
% % % 		\subcaption{\hspace*{2.2cm}(c) IMDB-B}
% % 		\end{minipage}
% % 	}
% 	\vskip -0.1in
% 	\caption{Records of .}
% 	\label{fig:visual}
% 	\vskip -0.1in
% \end{figure*}

% \subsection{Structural Analysis}
% \label{exp-sa}
% CLGL adopts the representation of the $m$-th layer of GNN to calculate $\mathcal{L}_a$. In order to prove the rationality of this design, we conducted multiple parameter experiments. We train on different datasets with different values of $m$ and report the performance of the model in Figure \ref{fig:m}. We can observe that on Spurious-Motif and SPMotif-Variant, the model performance is optimal when $m=2$, while on Graph-SST5, the model is optimal when $m=3$. Such a phenomenon justifies the necessity of having $m$ as a hyperparameter. Furthermore, it also proves that the GNN model tends to process and distinguishes different types of logic at different neural network layers.

\subsection{In-Depth Study (RQ2)}
\label{exp-va}

To understand how our model changes during training, we tracked accuracy and $\mathcal{L}_{a}$ loss across 200 epochs on multiple Motif-Variant datasets with different biases. $\mathcal{L}_{a}$ for ERM was calculated similarly to CLGL but without backpropagation. Results in Figure \ref{fig:ida} show that CLGL's $\mathcal{L}_{a}$ decreases quickly and stays consistently low, while its performance keeps improving even after multiple epochs. This suggests that CLGL can effectively learn and maintain the correct logic without degradation. Different biases had little impact on $\mathcal{L}_{a}$, indicating that the expertise logic isn't closely tied to downstream tasks. Therefore, its learning efficiency isn't affected by dataset variations.

% Figure \ref{fig:idb} displays the visualized m1del output features. Each horizontal line in each block in Figure \ref{fig:idb} represents the visualization of the model's representation of $\widetilde{X}$, and the vertical axis represents different samples. All samples within Figure \ref{fig:idb} come from same class. Therefore, the more consistent the vertical lines in the block, the more consistent the model's representation of causal information within the graph data across samples. Figure \ref{fig:idb} illustrates the specific changes in features during the learning process We can observe that CLGL produces more consistent representations at 100 and 200 rounds compared to ERM, which suggests that the logic is better learned by CLGL. Moreover, as seen in the figure, there are significant changes in the representations learned by CLGL between the 100th and 200th epochs, suggesting that it continues to learn over time. Figure \ref{fig:idc} displays different methods' feature representations of test samples at 200 epochs, indicating that CLGL offers a more refined and accurate representation.

Figure \ref{fig:idb} shows visualized output features of our model. Each horizontal line in each block represents the model's representation of graph data, with the vertical axis representing different samples from the same class. Consistent vertical lines indicate a more consistent representation of causal information within the graph data across samples. CLGL produces more consistent representations at the 100th and 200th epochs compared to ERM, indicating better expertise logic learning. Additionally, there are significant changes in CLGL's representations between the 100th and 200th epochs, indicating the model is continually learning. Figure \ref{fig:idc} displays different methods' feature representations of test samples at the 200th epoch, revealing that CLGL offers a more refined and accurate representation.

Figure \ref{fig:idd} demonstrates the impact of different hyperparameters on the model. The experimental results regarding $m$ and $K$ substantiate that extracting representations in the intermediate layers for causal knowledge learning, while simultaneously employing multiple interchange interventions, effectively enhances the model's performance. And, $\lambda$ can aid in striking a balance between causal learning and dataset training. These results validate the necessity of our proposed structures.

\section{Conclusions}

% By observing the motivating experiments, we derive an empirical conclusion that GNNs generally learn human expertise during training, and introducing expertise logic into graph representation learning can improve the model performance. Therefore, we introduce CLGL to lead the model to learn the expertise logic. We provide a theoretical analysis that substantiates the validity and effectiveness of our proposed design, highlighting its key strengths and justifying its implementation. Empirically, various comparisons prove that the proposed approach can yield significant performance boosts.

Through empirical observations of motivating experiments, we find that GNNs tend to learn human expertise during training, and introducing expertise logic into graph representation learning improves the model performance. Hence, we propose CLGL, a method that guides the model to learn expertise logic. The validity and effectiveness of our design are supported by theoretical analysis, highlighting its key strengths and justification. Empirical comparisons demonstrate the significant performance improvements achieved by our approach.

{\bf{Limitations}}. Our work is based on experimental observations to propose a novel research question and provide a solution. However, there is room for improvement in the optimization of the specific training process. Furthermore, the acquisition of expertise logic, which we utilized, actually requires extensive analytical work. Therefore, we will consider developing methods to automatically extract expertise logic from datasets in the future.

% By observing the motivating experiments, we derive the empirical conclusions that introducing expertise logic into graph representation learning can improve the model performance. Therefore, we introduce CLGL to lead the model to learn the expertise logic. We theoretically and empirically demonstrate the proposed approach.

% References follow the acknowledgments in the camera-ready paper. Use unnumbered first-level heading for
% the references. Any choice of citation style is acceptable as long as you are
% consistent. It is permissible to reduce the font size to \verb+small+ (9 point)
% when listing the references.
% Note that the Reference section does not count towards the page limit.
% \medskip

% {
% \small

% [1] Alexander, J.A.\ \& Mozer, M.C.\ (1995) Template-based algorithms for
% connectionist rule extraction. In G.\ Tesauro, D.S.\ Touretzky and T.K.\ Leen
% (eds.), {\it Advances in Neural Information Processing Systems 7},
% pp.\ 609--616. Cambridge, MA: MIT Press.

% [2] Bower, J.M.\ \& Beeman, D.\ (1995) {\it The Book of GENESIS: Exploring
%   Realistic Neural Models with the GEneral NEural SImulation System.}  New York:
% TELOS/Springer--Verlag.

% [3] Hasselmo, M.E., Schnell, E.\ \& Barkai, E.\ (1995) Dynamics of learning and
% recall at excitatory recurrent synapses and cholinergic modulation in rat
% hippocampal region CA3. {\it Journal of Neuroscience} {\bf 15}(7):5249-5262.
% }

\bibliographystyle{splncs04}
\bibliography{example_paper}

%%%%%%%%%%%%%%%%%%%%%%%%%%%%%%%%%%%%%%%%%%%%%%%%%%%%%%%%%%%%

\clearpage
\appendix

% \title{Appendix}
% \maketitle

\section{Proofs}
\subsection{A Validity Justification For The SCM In Figure \ref{fig:SCM}}
\label{apx:SCM}

To establish the validity of the proposed SCM in Figure \ref{fig:SCM}, we employ the IC algorithm \cite{DBLP:journals/ijon/Shanmugam01} to construct the SCM from scratch and provide the detailed construction process. IC algorithm is a method for identifying causal relationships from the ob
+served data. Please refer to Chapter 2 of \cite{DBLP:journals/ijon/Shanmugam01} for the details of the IC algorithm. 

The input of the IC algorithm is a set of variables and their distributions. The output is a pattern that represents the underlining causal relationships, which can be a structural causal model. The IC algorithm can be divided into three steps:

\paragraph{Step 1.} For each pair of variables $a$ and $b$ in $V$, the IC algorithm searches for a set $S_{ab}$ s.t. the conditional independence relationship $(a \upmodels b | S_{ab})$ holds. In other words, $a$ and $b$ should be independent given $S_{ab}$. The algorithm constructs an undirected graph $G$ with vertices corresponding to variables in $V$. A pair of vertices $a$ and $b$ are connected with an undirected edge in $G$ if and only if no set $S_{ab}$ can be found that satisfies the conditional independence relationship $(a \upmodels b | S_{ab})$.

\paragraph{Step 2.} For each pair of non-adjacent variables $a$ and $b$ that share a common neighbor $c$, check the existence of $c \in S_{ab}$. If such a $c$ exists, proceed to the next pair; if not, add directed edges from $a$ to $c$ and from $b$ to $c$.

\paragraph{Step 3.} In the partially directed graph results, orient as many of the undirected edges as possible subject to the following two conditions: (i) any alternative orientation of an undirected edge would result in a new y-structure, and (ii) any alternative orientation of an undirected edge would result in a directed cycle. 

Accordingly, we employ IC algorithm to provide a step-by-step procedure for constructing our structural causal model.

\begin{figure}[ht]
	\centering
	\subfigure[]{
		\begin{minipage}{0.22\textwidth} 
			\includegraphics[width=\textwidth]{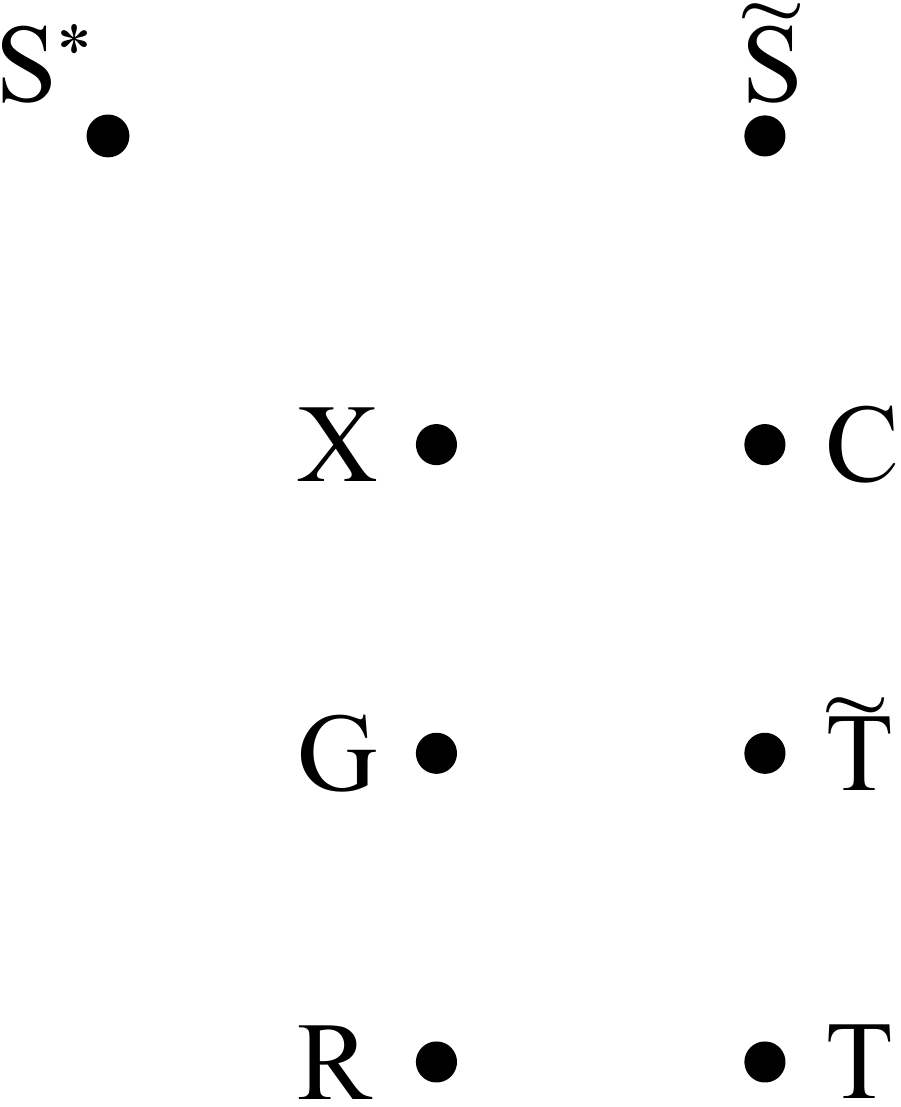} \\
                \label{IC-a}
		\end{minipage}
    }\hspace{2mm}
	\subfigure[]{
		\begin{minipage}{0.22\textwidth} 
			\includegraphics[width=\textwidth]{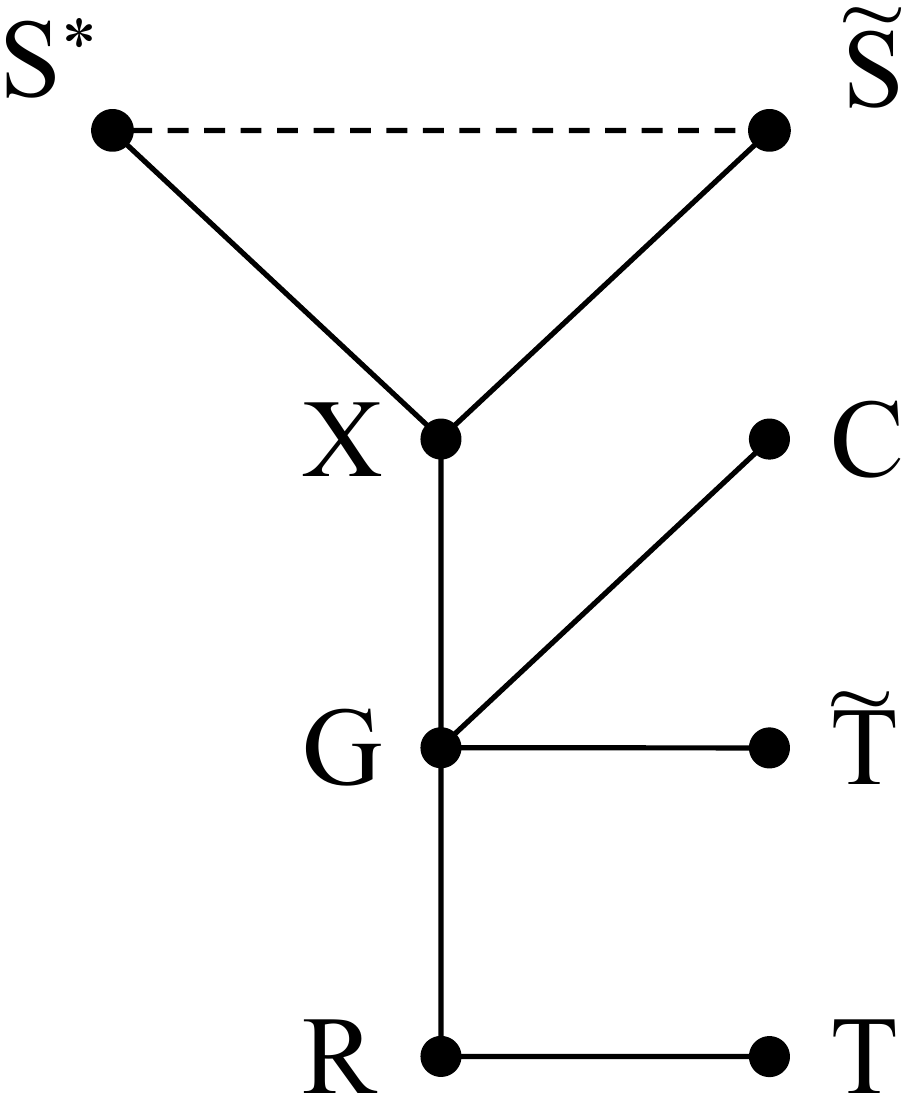} \\
                \label{IC-b}
		\end{minipage}
	}\hspace{2mm}
 	\subfigure[]{
		\begin{minipage}{0.22\textwidth} 
			\includegraphics[width=\textwidth]{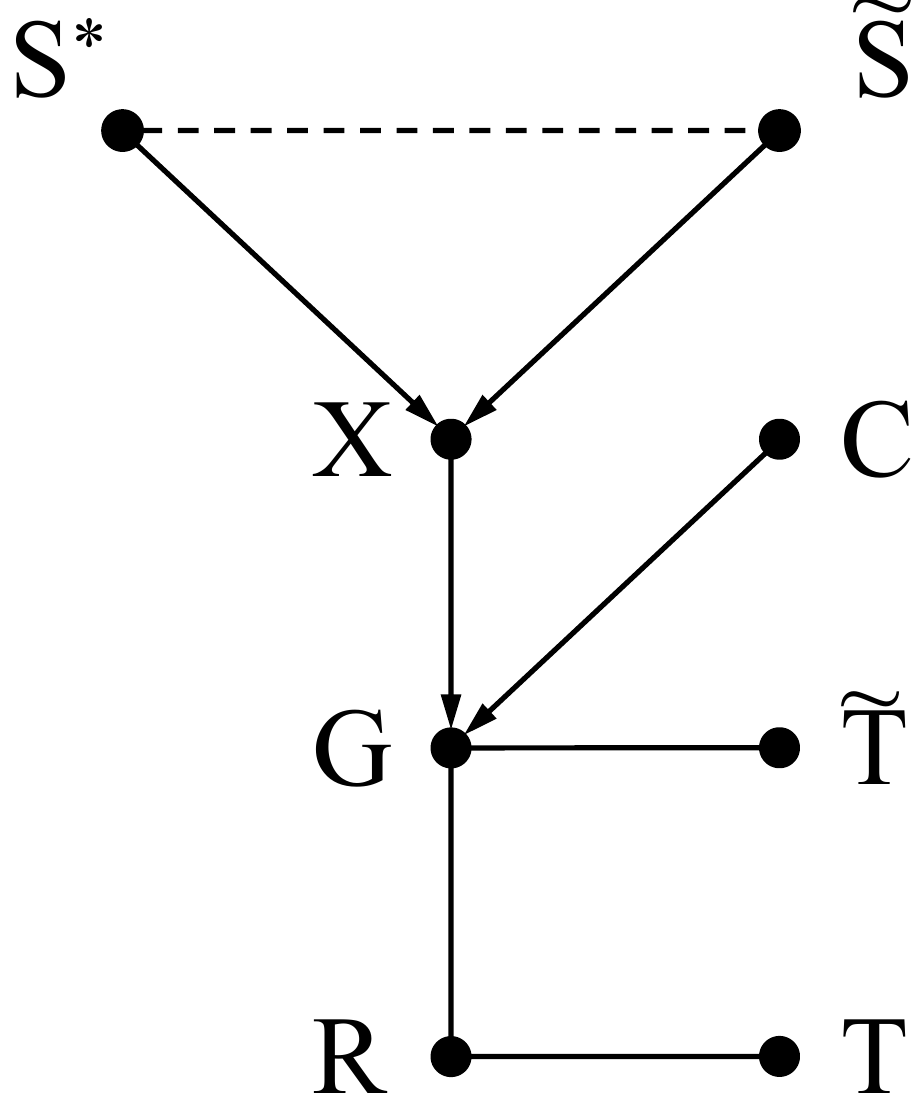} \\
                \label{IC-c}
		\end{minipage}
	}\hspace{2mm}
  	\subfigure[]{
		\begin{minipage}{0.22\textwidth} 
			\includegraphics[width=\textwidth]{IC-d.pdf} \\
                \label{IC-d}
		\end{minipage}
	}

	% \vskip -0.05in
	\caption{Visualization of the reasoning process using IC algorithm.}
	\label{fig:IC}
	%\vskip -0.1in
\end{figure}

As for step 1, we first represent all variables as nodes in Figure \ref{IC-a}. For each node, we traverse all other nodes to determine whether to establish a connection. Accordingly, $R$ represents the model's representation of $G$. Since any other variable can only affect $R$ through $G$, $R$ is conditionally independent of all other variables given $G$. Therefore, $R$ is only connected to $G$. $G$ represents the graph data, which is composed of $X$ and $C$. There is no other variable that can block the relationship between $G$ and $X$, or $G$ and $C$. Therefore, $G$ is connected to both $X$ and $C$. As $X$ blocks the path from $S^*$ and $\widetilde{S}$ to $G$, the corresponding connections do not exist. We define $C$ as a confounder in the graph data, which has no causal relationship with other variables. Therefore, an empty set can make $C$ independent of other variables, except for $G$.

As $\widetilde{S}$ and $S^{*}$ are all graph causal factors, each of them is correlated with $X$. Their connections cannot be blocked by any set of variables, and thus we connect all these nodes with $X$. The relationship between $S^{*}$ and $\widetilde{S}$ can not be figured out, therefore we adopt a dashed line to link them. The result is demonstrated in Figure \ref{IC-b}.

We then move on to step 2. Similar to step 1, we process with the traversal analysis starting from $R$. For $R$, since $G \in S_{RG}$ and $G \in S_{RX}$, no edges related to $R$ can be directed. Then, as $S_{XC} = \emptyset$, $X \notin S_{XC}$, we direct edge $X$ to $G$ and $C$ to $G$. As $S^{*}$ and $\widetilde{S}$ is the cause of $X$ by definition, therefore we direct edge $S^{*}$ to $X$ and $\widetilde{S}$ to $X$.

% As $S^{*}$ is independent of $S'$ and $\widetilde{S}$ by definition, $S_{S^{*}\widetilde{S}} = \emptyset$ and $S_{S^{*}S'} = \emptyset$. We then direct edge $S^{*}$ to $X$ and $\widetilde{S}$ to $X$. The result of step 2 are demonstrated in Figure \ref{IC-c}. 

In Step 3, we adopted the rule 1 from \cite{DBLP:conf/uai/VermaP92} for systematizing this step, which states that if $a \to b$ and $a$ and $c$ are not adjacent, then $b \to c$ should be set as the orientation for $b - c$. Based on this, we established the following orientations: $G \to \widetilde{T}, G \to R$. Then, according to the same rule, $R \to T$ should also be set. As for the remaining edge $(\widetilde{S},S^{*})$, since we cannot determine its direction, we represent this edge with a bidirectional dashed line, indicating its directionality is uncertain. The final result is illustrated in Figure \ref{IC-d}, which is identical to the SCM in Figure \ref{fig:SCM}.

% we  does not assist us in orienting any edges. However, for the edge $(R, G)$, since $R$ is determined by the neural network model based on $G$, i.e., $R = f(G)$, we can create an edge from $G$ to $R$. As for the remaining edge $(S,S')$, since we cannot determine its direction, we represent this edge with a bidirectional dashed line, indicating its directionality is uncertain. The final result is illustrated in Figure \ref{IC-d}, which is identical to the SCM in Figure \ref{fig:SCM}.

\subsection{Proof of Theorem \ref{th:c}}
\label{apx:pfthc}
To prove the theorem, we follow \cite{pearl2009causality} and suppose the proposed SCM possesses Markov property. Therefore, according to the SCM in Figure \ref{fig:SCM}, $C$ and $R$ are conditionally independent given $G$, as $G$ blocks any path between $C$ and $R$. Therefore, we could apply the Data Processing Inequality \cite{DBLP:books/wi/01/CT2001} on the path $C \to G \to R$. Formally, we have:
\begin{gather}
I(G;C) \geq I(R;C).
\end{gather} 
According to the Chain Rule for Information \cite{DBLP:books/wi/01/CT2001}, we also have:
\begin{gather}
I(X,C;G) = I(X;G|C) + I(C;G).
\end{gather} 
As $X$ and $C$ are independent, we have:
\begin{gather}
I(X,C;G) = I(X;G) + I(C;G).
\end{gather} 
Therefore:
\begin{gather}
 I(C;G) = I(X,C;G) - I(X;G).
\end{gather} 
As $G$ is the graph data that consist of $X$ and $C$, we have:
\begin{gather}
 I(X,C;G) = 1.
\end{gather} 
Then:
\begin{gather}
 I(C;G) = 1 - I(X;G).
\end{gather} 
Therefore:
\begin{gather}
\label{eq:1-I}
1 - I(X;G) \geq I(R;C).
\end{gather} 
According to the SCM in Figure \ref{fig:SCM}, $\widetilde{S}$ and $R$ are conditionally independent given $G$, we have:
\begin{gather}
\label{eq:ptc_sgsr}
I(\widetilde{S};G) \geq I(\widetilde{S};R).
\end{gather} 
Also:
\begin{gather}
I(\widetilde{S},X;G) = I(\widetilde{S};G) + I(X;G|\widetilde{S})
\end{gather} 
and
\begin{gather}
I(\widetilde{S},X;G) = I(X;G) + I(\widetilde{S};G|X)
\end{gather} 
holds. $\widetilde{S}$ and $G$ is independent given $X$, therefore:
\begin{gather}
I(\widetilde{S};G|X) = 0.
\end{gather} 
Thus:
\begin{gather}
I(\widetilde{S},X;G) = I(\widetilde{S};G) + I(X;G|\widetilde{S}) = I(X;G).
\end{gather} 
As $I(X;G|\widetilde{S}) \geq 0$:
\begin{gather}
\label{eq:ptc_sgxg}
I(\widetilde{S};G) \leq I(X;G).
\end{gather} 
Based on Inequality \ref{eq:ptc_sgsr} and \ref{eq:ptc_sgxg}, we have:
\begin{gather}
\label{eq:ptc_sfinall}
I(\widetilde{S};R) \leq I(X;G).
\end{gather} 
Substituting Inequality \ref{eq:ptc_sfinall} into Inequality \ref{eq:1-I}, we can obtain:
\begin{gather}
I(R;C) \leq 1-I(\widetilde{S};R).
\end{gather} 
The theorem is proved.

\subsection{Proof of Theorem \ref{th:ori}}
\label{apx:pfori}
We begin with calculating the boundaries of $I(R,\widetilde{S})$. According to the SCM in Figure \ref{fig:SCM}, $\widetilde{S}$ and $G$ are conditionally independent given $X$, as $X$ block any path between $\widetilde{S}$ and $G$. Therefore, we could apply the Data Processing Inequality \cite{DBLP:books/wi/01/CT2001} on the path $\widetilde{S} \to X \to G$. Formally, we have:
\begin{gather}
I(\widetilde{S}; X) \geq I(\widetilde{S};G).
\end{gather} 
Likewise, we have:
\begin{gather}
I(\widetilde{S}; G) \geq I(\widetilde{S};R),
\end{gather} 
and:
\begin{gather}
I(\widetilde{S}; R) \geq I(\widetilde{S};T).
\end{gather} 
According to the assumptions:
\begin{gather}
I(\widetilde{T}; \widetilde{S}) = I(G;\widetilde{S}),
\end{gather} 
we have:
\begin{gather}
I(\widetilde{S} ; \widetilde{T}) \geq I(\widetilde{S};R) \geq I(\widetilde{S};T).
\end{gather} 
So far, we can acquire the upper and lower bounds of $I(\widetilde{S};R)$. Moreover, since the data set and the expert system are determined in advance in the learning task, the only thing that can be changed for this system is the parameters of the neural network model. Therefore, the upper bound $I(\widetilde{S} ; \widetilde{T})$ holds a fixed value. If we can make the lower bound $I(\widetilde{S};T)$ equal to the upper bound $I(\widetilde{S} ; \widetilde{T})$, then we have $I(\widetilde{S};R)$ reached the maximum. Next, we will proof if $p(t|\widetilde{s}) = q(t|\widetilde{s})$, then $I(\widetilde{S} ; \widetilde{T}) = I(\widetilde{S} ; T)$.

When $p(t|\widetilde{s}) = q(t|\widetilde{s})$, then: 
\begin{align}
H(T|\widetilde{S}) &= - \sum_{\widetilde{s} \in \mathcal{\widetilde{S}}} p_{\widetilde{S}}(\widetilde{s})\sum_{ t \in \mathcal{T}} p(t| \widetilde{s}) log\Big( p(t| \widetilde{s})\Big) \nonumber\\
&=  - \sum_{\widetilde{s} \in \mathcal{\widetilde{S}}} p_{\widetilde{S}}(\widetilde{s})\sum_{ t \in \mathcal{T}} q(t| \widetilde{s}) log\Big( q(t| \widetilde{s})\Big) \nonumber\\
&= H(\widetilde{T}|\widetilde{S}),
\label{TSeq} 
\end{align}
where $\widetilde{S} \sim p_{\widetilde{S}}(\widetilde{s}), \widetilde{s} \in \mathcal{\widetilde{S}}$.

We also have:
\begin{align}
H(T) &=  - \sum_{t \in \mathcal{T}} p(t) log\Big( p(t) \Big) \nonumber\\
&=  - \sum_{t \in \mathcal{T}} \sum_{\widetilde{s} \in \mathcal{\widetilde{S}}}p_{\widetilde{S}}(\widetilde{s})p(t|\widetilde{s}) log\Big( \sum_{\widetilde{s} \in \mathcal{\widetilde{S}}}p_{\widetilde{S}}(\widetilde{s})p(t|\widetilde{s}) \Big) \nonumber\\
&=  - \sum_{t \in \mathcal{T}} \sum_{\widetilde{s} \in \mathcal{\widetilde{S}}}p_{\widetilde{S}}(\widetilde{s})q(t|\widetilde{s}) log\Big( \sum_{\widetilde{s} \in \mathcal{\widetilde{S}}}p_{\widetilde{S}}(\widetilde{s})q(t|\widetilde{s}) \Big) \nonumber\\
&=    - \sum_{t \in \mathcal{T}} q(t) log\Big( q(t) \Big) \nonumber\\
&= H(\widetilde{T})
\label{Teq} 
\end{align}

% \begin{gather} 
% \frac{P(\widetilde{T},\widetilde{S})}{P(\widetilde{S})} = \frac{P(T,\widetilde{S})}{P(\widetilde{S})}, \\
% P(\widetilde{T},\widetilde{S}) = P(T,\widetilde{S}).\label{TSeq} 
% \end{gather} 
% If we denote all the event within $\widetilde{S}$ as $\widetilde{S}_n:n = 1,2,3,...,n$, then according to the law of total probability, we have: 
% \begin{gather}
% P(\widetilde{T})=\sum_{i=0}^{n}P(S_i)P(\widetilde{T}|S_i), 
% \end{gather} 
% and,
% \begin{gather}
% P(T)=\sum_{i=0}^{n}P(S_i)P(T|S_i),
% \end{gather} 
% As $P(\widetilde{T}|\widetilde{S}) = P(T|\widetilde{S})$, then:
% \begin{gather}
% P(T)=P(\widetilde{T}).
% \label{Teq}
% \end{gather} 
% Based on Equation \ref{Teq} and \ref{TSeq}, we have:
% \begin{gather}
% \frac{P(\widetilde{T},\widetilde{S})}{P(\widetilde{T})} = \frac{P(T,\widetilde{S})}{P(T)}, \\
% P(\widetilde{S}|\widetilde{T}) = P(\widetilde{S}|T).
% \end{gather} 
% Therefore:
% \begin{gather}
% H(\widetilde{S}|\widetilde{T}) = H(\widetilde{S}|T).
% \end{gather} 

According to the definition of mutual information \cite{DBLP:books/wi/01/CT2001}, we have:
\begin{gather}
I(\widetilde{S} ; \widetilde{T}) = H(\widetilde{T}) - H(\widetilde{T} | \widetilde{S}), \nonumber\\
I(T ; \widetilde{S}) = H(T) - H(T | \widetilde{S}).
\label{IHH}
\end{gather} 
Based on equation \ref{TSeq}, \ref{Teq} and \ref{IHH}, we have:
\begin{gather}
I(\widetilde{S} ; \widetilde{T}) = I(\widetilde{S} ; T).
\end{gather} 
Based on the discussions above, $I(\widetilde{S};R)$ reached the maximum. The theorem is proved.

\subsection{Proof of Corollary \ref{corollary:Lc2}}
\label{apx:pfc}
As in Equation \ref{eq:Lo}, for each graph $G_{i}$, $\widetilde{S}$ holds a fixed value $\widetilde{s}_{i}$. Therefore, we have:
\begin{gather}
p_{i}(t) = p_{i}(t|\widetilde{s}_{i}) = p_{i}(t|\widetilde{s}),
\end{gather} 
and 
\begin{gather}
\hat{q}_{i}(t) = \hat{q}_{i}(t|\widetilde{s}_{i}) = \hat{q}_{i}(t|\widetilde{s}).
\end{gather} 
Therefore, we have:
\begin{gather}
\mathcal{L}_{c} = \sum_{i}^{N} KL\Big(p_{i}(t|\widetilde{s}) || \hat{q}_{i}(t|\widetilde{s})\Big) = \sum_{i}^{N} KL\Big(p_{i}(t) || \hat{q}_{i}(t)\Big)
\end{gather} 
As $r^{P}(r^{E}(G_{i}))$ can predict $p_{i}(t)$. $f^{P}(f^{E}(G_{i}))$ output a estimation $\hat{q}_{i}(t)$ of ${q}_{i}(t)$. Therefore:

\begin{gather}
\mathcal{L}_{c} = \sum_{i}^{N} KL\Big(p_{i}(t) || \hat{q}_{i}(t)\Big) = \sum_{i=1}^{N} \mathcal{D}_{KL}\Big(r^{P}(r^{E}(G_i)), f^{P}(f^{E}(G_i))\Big)
\end{gather} 

Then, if:

\begin{gather}
\mathcal{L}_{c} =  \sum_{i=1}^{N} \mathcal{D}_{KL}\Big(r^{P}(r^{E}(G_i)), f^{P}(f^{E}(G_i))\Big) = 0,
\end{gather} 

we have:

\begin{gather}
\sum_{i}^{N} KL\Big(p_{i}(t) || \hat{q}_{i}(t)\Big) = \sum_{i}^{N} KL\Big(p_{i}(t|\widetilde{s}) || \hat{q}_{i}(t|\widetilde{s})\Big) = 0.
\end{gather} 

Further more, the training samples in $G$ sufficiently cover all embodiments corresponding to $\widetilde{S}$, we have:

\begin{gather}
KL\Big(p(t|\widetilde{s}) || q(t|\widetilde{s})\Big) = 0.
\end{gather} 

Then $p(t|\widetilde{s})$ equals $q(t|\widetilde{s})$, according to Theorem \ref{th:ori}, $I(R;\widetilde{S})$ reaches maximization. The Corollary is proved.

\section{Experiment Setting Details.}
\label{app:set}
In this section, we give a detailed description of our experiment settings. 

\begin{table*}[h] \small
	\centering
 	\caption{Summary of datasets.}
	\begin{tabular}{lccccccc}
		\toprule
		Name      &  Graphs\# & Average Nodes\# & Classes\# & Task Type &Metric       \\
		\midrule
		Spurious-Motif &18,000 &46.6 &3 &Classification &ACC \\
		Motif-Variant  &18,000 &48.9 &3 &Classification &ACC \\
            Graph-SST5 &11,550 &21.1 &5 &Classification &ACC \\
		Graph-Twitter &6,940 &19.8 &3 &Classification &ACC \\
		\bottomrule
		
	\end{tabular}

	\label{tab:datasum}
\end{table*}

\begin{table*}[h] \small
	\centering
 	\caption{Summary of the backbones used in each dataset.}
	\begin{tabular}{lccccc}
		\toprule
		Dataset  & Backbone & GNN structure   & Global Pool       \\
		\midrule
		Spurious-Motif &Local Extremum GNN &[4,32,32,32]  & global mean pool  \\
            Motif-Variant &Local Extremum GNN &[4,32,32,32] & global mean pool  \\
		Graph-SST5 &ARMA &[768,128,128]  & global mean pool  \\
		Graph-Twitter &ARMA &[768,128,128]  & global mean pool  \\
		\bottomrule
	\end{tabular}

	\label{tab:network}
\end{table*}

\subsection{Datasets}
\label{apx:dataset}
We evaluate our method on four different datasets, and the ID/OOD version of these datasets. The datasets are built upon manually constructed data and sentiment graph data. We summarize the datasets we used in Table \ref{tab:datasum}. Furthermore, we demonstrate the backbone GNN we adopt for each dataset in Table \ref{tab:network}. The GNNs we utilized included Local Extremum GNN \cite{DBLP:conf/aaai/RanjanST20}, ARMA \cite{DBLP:conf/aaai/0001RFHLRG19}.

\textbf{Spurious-Motif}. Spurious-Motif is an OOD synthetic dataset proposed by \cite{DBLP:conf/nips/YingBYZL19}. We adopt a reimplemented version created by \cite{DBLP:conf/iclr/WuWZ0C22}. The dataset involves 18,000 graphs, each one of them consisting of two subgraphs: motif and confounder. The motif consists of ground truth data with fixed structures and is causally related to the ground truth label. Confounder, on the other hand, has no causal relationship with the labels. Spurious-Motif possessed three types of motifs and confounders. In the training set, confounders are stitched into the graph data with certain biases. For example, a bias of 0.9 means that for each class, 90$\%$ of motifs and confounders that are stitched together belong to the same class. The motifs and confounders in the test set are appended randomly. Therefore, the greater the bias, the harder it is for the model to distinguish between causal information and confounders. Additionally, the test graph data includes large subgraphs to make the learning task harder.

% Spurious-Motif is a synthetic dataset created by \cite{ying2019gnnexplainer}. We adopt the re-implementation version created by \cite{DBLP:conf/iclr/WuWZ0C22}, which involves 18, 000 graphs. Each graph in the dataset consists of two subgraphs. One serves as the ground-truth data that is causally related to the graph label, the other serves as the confounder. There are three types of ground truth and confounder subgraphs each. In the training set, the confounder subgraph was stitched to the graph data with a certain bias. e.g., if the bias is 0.7, that means for each class, 70$\%$ of the samples are stitched with the same kind of confounder subgraph. The ground-truth subgraphs and confounder subgraphs in the test sets are randomly attached. Therefore, the higher the bias, the harder for the model to distinguish the causal information from confounders. Additionally, the dataset includes graphs with large subgraphs to make the learning task harder. 

\textbf{Motif-Variant}. As the causal information in Spurious-Motif consists of motifs with fixed structures, we modify the dataset to have more varied data. Instead of fixed structures, we adopt different kinds of motifs and allow some shape changes on them. Based on such changes, we are able to create a larger gap between the training and the test set.

\textbf{Graph-SST5}. Graph-SST5 dataset \cite{yuan2020explainability} is an ID sentiment graph dataset that builds upon the Rotten Tomatoes comments. 

\textbf{Graph-Twitter}. Graph-Twitter dataset \cite{yuan2020explainability} is also a sentiment graph dataset. However, its data comes from the comments on Twitter.

\textbf{Spurious-Motif (ID)}. The ID version of Spurious-Motif, which removes the bias and the large subgraphs in test data.

\textbf{Motif-Variant (ID)}. The ID version of Motif-Variant, following the same pattern as Spurious-Motif (ID).

\textbf{Graph-SST5 (OOD)}. The OOD version of Graph-SST5. We follow \cite{DBLP:conf/iclr/WuWZ0C22} to split the graph samples into different sets according to their average node degree.

\textbf{Graph-Twitter (OOD)}. The OOD version of Graph-Twitter, in which we adopt the same modification pattern as Graph-SST5 (OOD).

% We follow \cite{DBLP:conf/iclr/WuWZ0C22} to split the graph samples into different sets according to their average node degree to increase the difficulty of the training.

% % \subsubsection{Datasets without artificial confounders.} These datasets consist of real-world graph data without artificial confounders. 

% (3) Graph-SST2 dataset (ID) \cite{yuan2020explainability}. The original Graph-SST2 dataset.

% (4) Graph-Twitter dataset \cite{yuan2020explainability}. Also, a sentiment graph dataset as Graph-SST2 dataset, but with different data sources.

% (5) Mol-BBBP \& Mol-BACE \cite{hu2020open}. Molecule datasets from OGBG datasets.

\subsection{Environments and Optimization Method}
% All our experiments were conducted on a workstation with two Quadro RTX 5000 GPU (16 GB), one Intel Xeon E5-1650 CPU, 128GB RAM, and an Ubuntu 20.04 operating system. During training, we adopt the Adam optimizer. We set the maximum training epoch as 300 for all tasks. For backpropagation, We use Stochastic Gradient Descent (SGD) for the optimization on Graph-SST2, Graph-Twitter, Mol-BBBP, and Mol-BACE, and Gradient Descent (GD) for the Spurious-Motif.

All our experiments are performed on a workstation equipped with two Quadro RTX 5000 GPUs (16 GB), an Intel Xeon E5-1650 CPU, 128GB RAM and Ubuntu 20.04 operating system. We employ the Adam optimizer for optimization. We set the maximum training epochs to 400 for all tasks. For backpropagation, we optimize Graph-SST2 and Graph-Twitter using stochastic gradient descent (SGD), while utilizing gradient descent (GD) on Spurious-Motif and Motif-Variant.

\subsection{Hyperparameters}
\label{app:hyp}

As for the hyperparameters, we set $m$ to 2 for the Spurious-Motif and Motif-Variant datasets, and 3 for the Graph-SST5 and Graph-Twitter datasets. We set $\lambda$ to 1 for the Spurious-Motif and Motif-Variant datasets and 1.5 for the Graph-SST5 and Graph-Twitter datasets. We set $K$ as 3 for all datasets. We set the learning rate as 0.001 for Spurious-Motif and Motif-Variant and 0.0002 for others. We train our model for 200 epochs, then stop training early if there is no performance improvement on the validation set for five epochs. We choose the model with the best validation performance as the final result. The maximum number of training epochs is set to 400 for all datasets. The training batch size is set as 32 for all datasets. The structure of GNN adopted for different datasets is shown in the table \ref{tab:network}.

\section{Implementation Details}
\label{apx:r}
To guide GNNs to learn expertise logic, we build multiple higher-order causal models corresponding to different kinds of expert logic. In this part, we will describe the detailed implementations of these models, along with other detailed implementations of CLGL.

\subsection{Implentation of $r^{E}(\cdot)$}
\label{apx:re}
\paragraph{$r^{E}(\cdot)$ concerning structure information} The expertise logic we employ is the "structural morphology of connections between subgraphs." As datasets rich in structural information often consist of multiple substructures with distinct features, the information regarding the structural morphology at the connections between these substructures becomes an expertise logic that can help guide the neural network learning more knowledge. Specifically, to enable the recognition of connection points by $r^{E}(\cdot)$, we first preprocess the graph data used for training and identify and label the connection points using topological structure matching techniques. For ease of training, we only label one connection point per graph sample, and this label also guides the training of $f^{P}(\cdot)$. Due to the reuse of training data, this representation will be precomputed in advance.

Due to our assumption of $I(\widetilde{T};\widetilde{S})=I(G;\widetilde{S})$ in Theorem \ref{th:ori}, we hereby provide an explanation for why the designed $r^E(\cdot)$ adheres to this assumption. We can assert that $\widetilde{S}$ in our discussion represents the analyzable and representable aspects of $r^E(\cdot)$ based on G. Thus, even if $G$ may contain information beyond $I(G;\widetilde{S})$, we can exclude it from the scope of $\widetilde{S}$.

% Subsequently, based on the graph structure, $r^{E}(\cdot)$ outputs a fixed score. 
% Since $r^{E}(\cdot)$'s input is a specific connection point and we do not perform further classification on this basis, the output is always a fixed value.

\paragraph{$r^{E}(\cdot)$ concerning node attribution} We have designed a second type of $r^{E}(\cdot)$ based on node attribute-related information. Since our target dataset is related to natural language, we utilize the concept of conjunctions, specifically the conjunction "but" and its representation of contrast relationships, as the expertise logic for designing $r^{E}(\cdot)$. Specifically, we preprocess the graph data used for training and identify the nodes corresponding to "but" and the nodes representing the respective sentences in which they appear. $r^{E}(\cdot)$ then outputs the degree of semantic contrast between the text preceding and following "but."

% Due to the difficulty of further classifying the degree of contrast, the output is also a fixed value.

\subsection{Implentation of $r^{I}(\cdot)$}
\label{apx:ri}
% As mentioned earlier, the outputs of the neural network $r^{E}(\cdot)$ we use are fixed values. 

In practice, due to the difficulty of finely modeling the composition of graph data in multiple cases, the expertise logic we can obtain contains rather generic knowledge. In such situations, the space of possible values for the graph causal factor may be small. Training an $r^{E}(\cdot)$ under such conditions would be ineffective. Therefore, we perform interchange intervention operations. Based on such operations, $r^{I}(\cdot)$ will introduce other information beyond embodiment to change its value. As demonstrated in Equation \ref{eq:ri}, $r^{I}(\cdot)$ generates different weights based on the type of interchange intervention. The specific type of interchange intervention performed directly determines the weights that $h(\cdot)$ outputs. Specifically, different interchange interventions can induce changes in the learned representation of the embodiment. We assess the magnitude of these changes and subsequently evaluate the probability that the post-intervention embodiment belongs to the same category as the pre-intervention embodiment. Based on this assessment, we generate the weights accordingly. 

\subsection{Implentation of $r^{P}(\cdot)$}
\label{apx:rp}
$r^{P}(\cdot)$ predicts the probabilities of different values for the graph causal factor based on the output of $r^{E}(\cdot)$. The predicted values are determined using a predefined formula, which would project the outputs of $r^{E}(\cdot)$ into a vector that represents the probability density function. Such a vector will then be used to calculate the KL divergence.

\begin{figure*}[ht]
	\centering
	\subfigure[]{
		\begin{minipage}{0.48\textwidth} 
			\includegraphics[width=\textwidth]{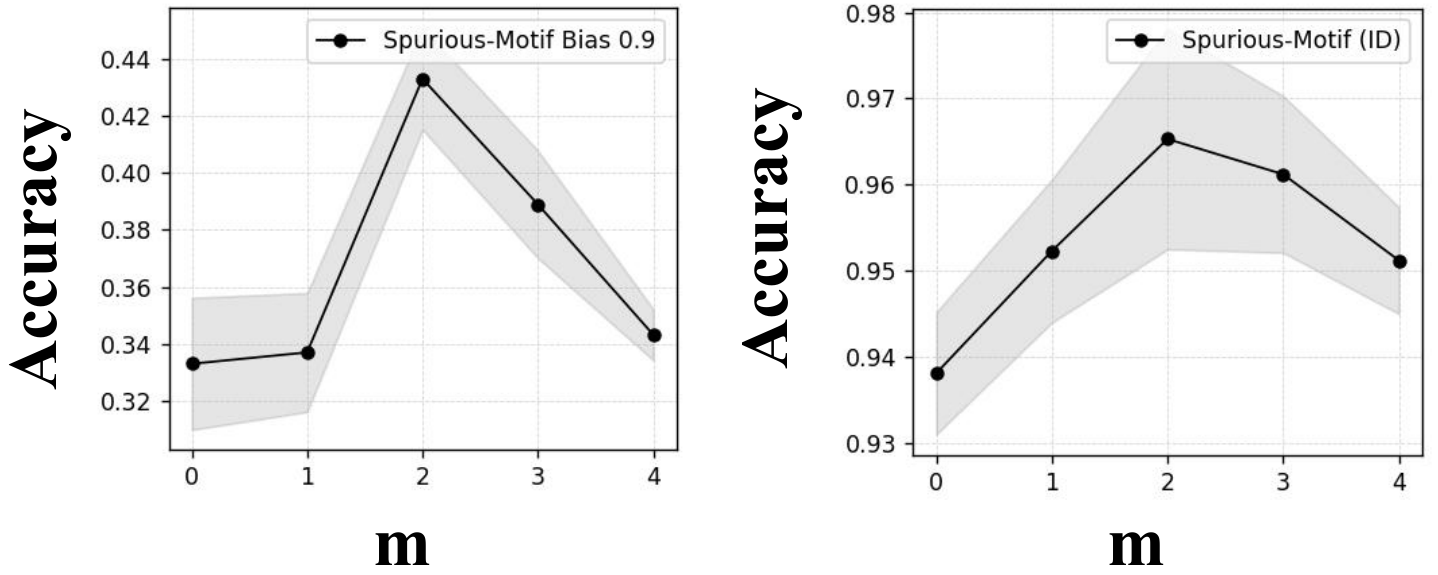} \\
%  		\subcaption{\hspace*{2.3cm}(a) MUTAG}
        \label{pic:m-a}
		\end{minipage}
	}
	\subfigure[]{
		\begin{minipage}{0.48\textwidth} 
			\includegraphics[width=\textwidth]{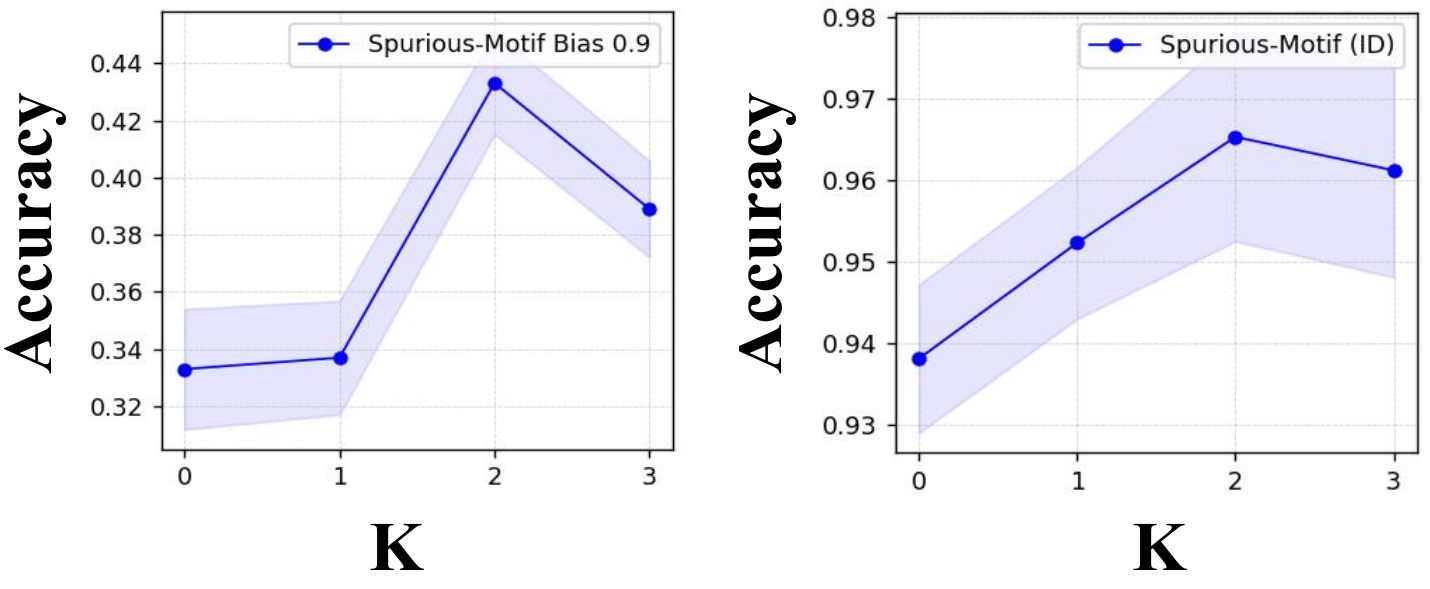} \\
%  		\subcaption{\hspace*{2.3cm}(a) MUTAG}
        \label{pic:m-b}
		\end{minipage}
	}
 	\subfigure[]{
		\begin{minipage}{0.48\textwidth} 
			\includegraphics[width=\textwidth]{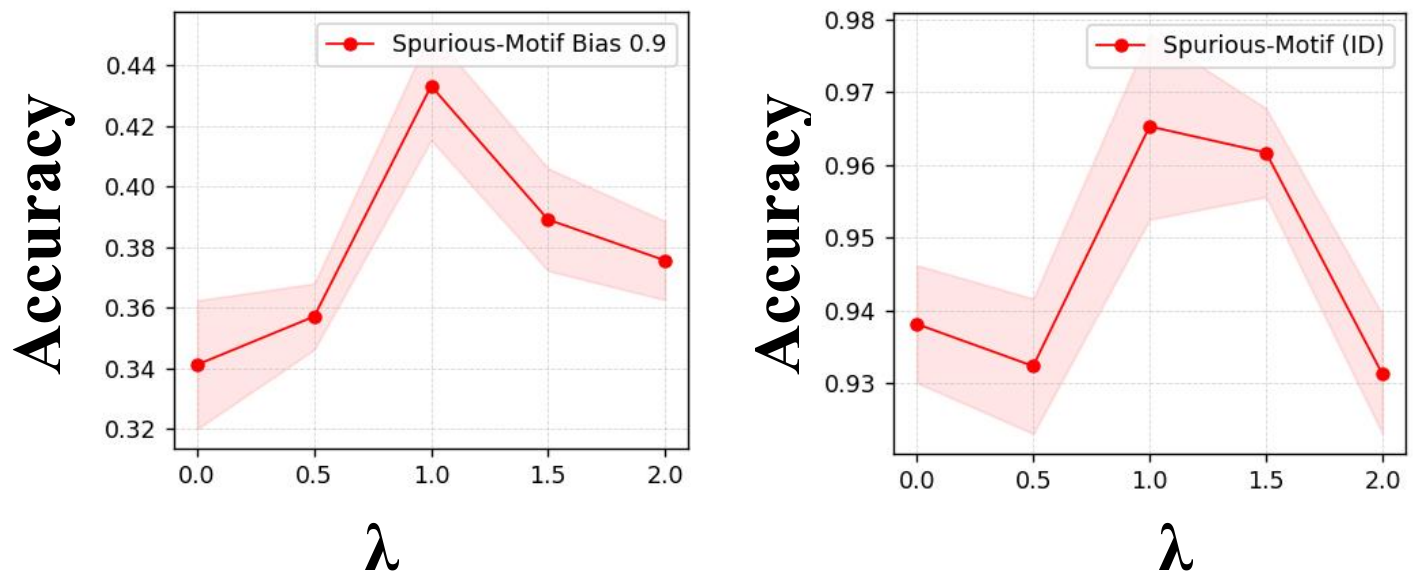} \\
%  		\subcaption{\hspace*{2.3cm}(a) MUTAG}
        \label{pic:m-c}
		\end{minipage}
	}

	\caption{Performance of CLGL on Spurious-Motif dataset with different hyperparameter values. The translucent region represents the confidence interval of the results.}
	\label{fig:spm}
	%\vskip -0.1in
\end{figure*}

\begin{figure*}[ht]
	\centering
	\subfigure[]{
		\begin{minipage}{0.45\textwidth} 
			\includegraphics[width=\textwidth]{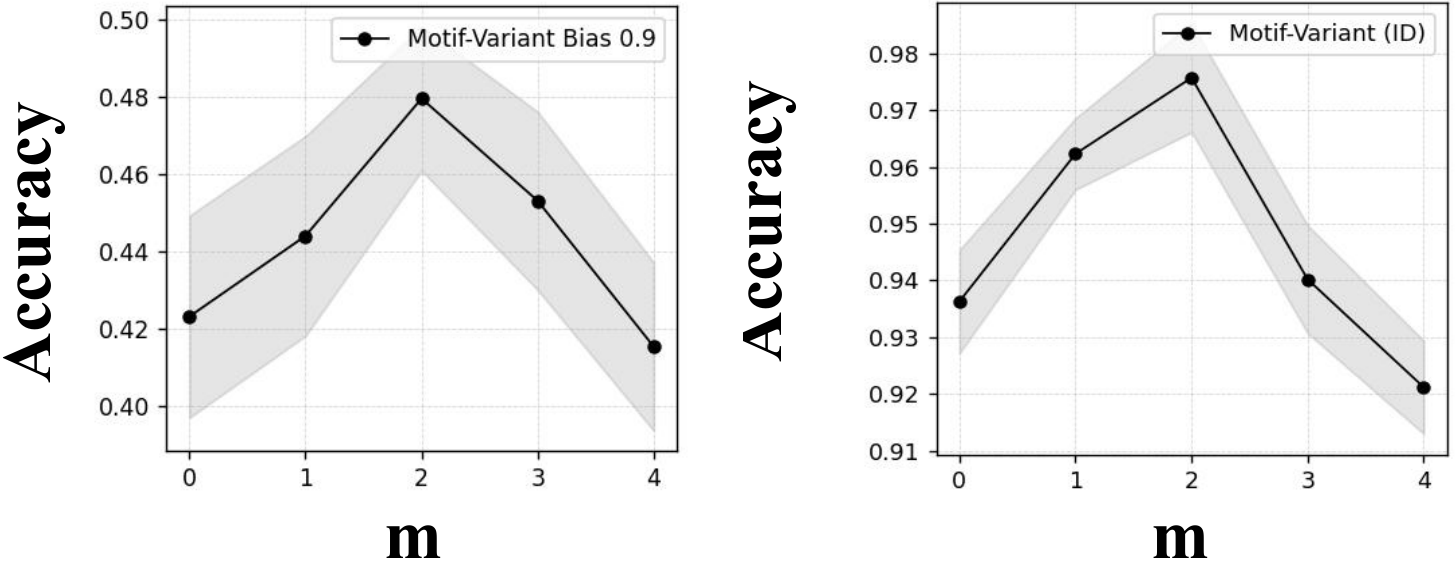} \\
%  		\subcaption{\hspace*{2.3cm}(a) MUTAG}
        \label{pic:m-a}
		\end{minipage}
	}\hspace{2mm} 
	\subfigure[]{
		\begin{minipage}{0.45\textwidth} 
			\includegraphics[width=\textwidth]{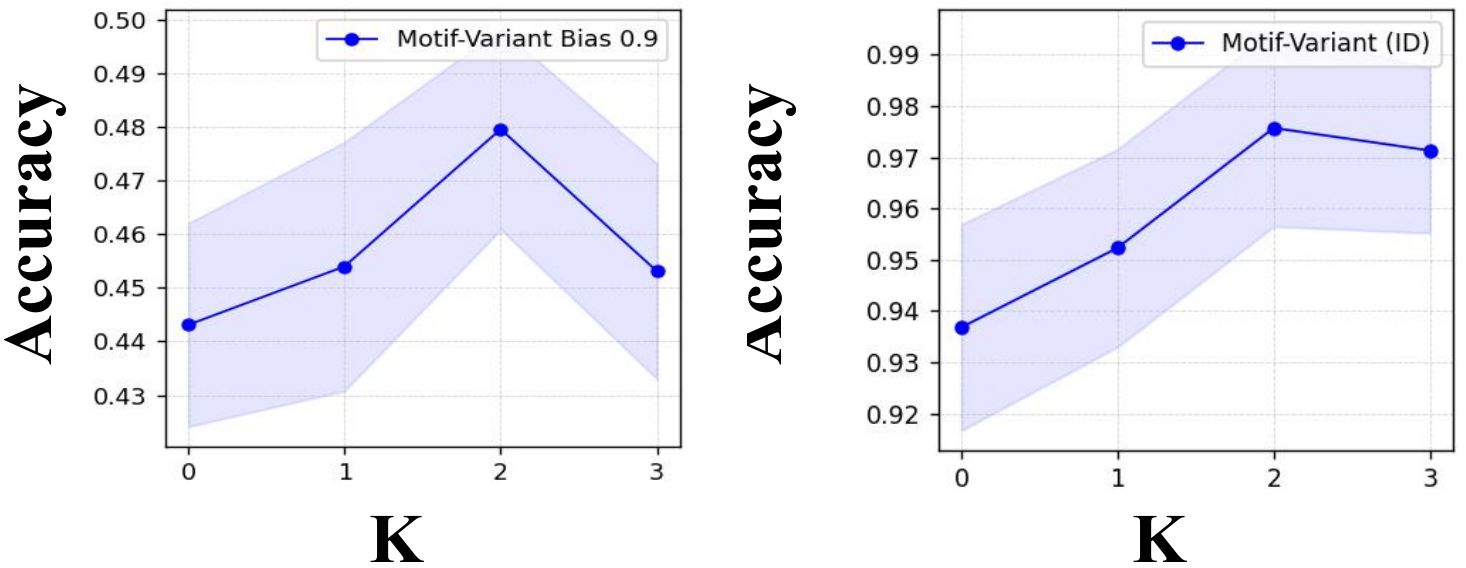} \\
%  		\subcaption{\hspace*{2.3cm}(a) MUTAG}
        \label{pic:m-b}
		\end{minipage}
	}\hspace{2mm} 
 	\subfigure[]{
		\begin{minipage}{0.45\textwidth} 
			\includegraphics[width=\textwidth]{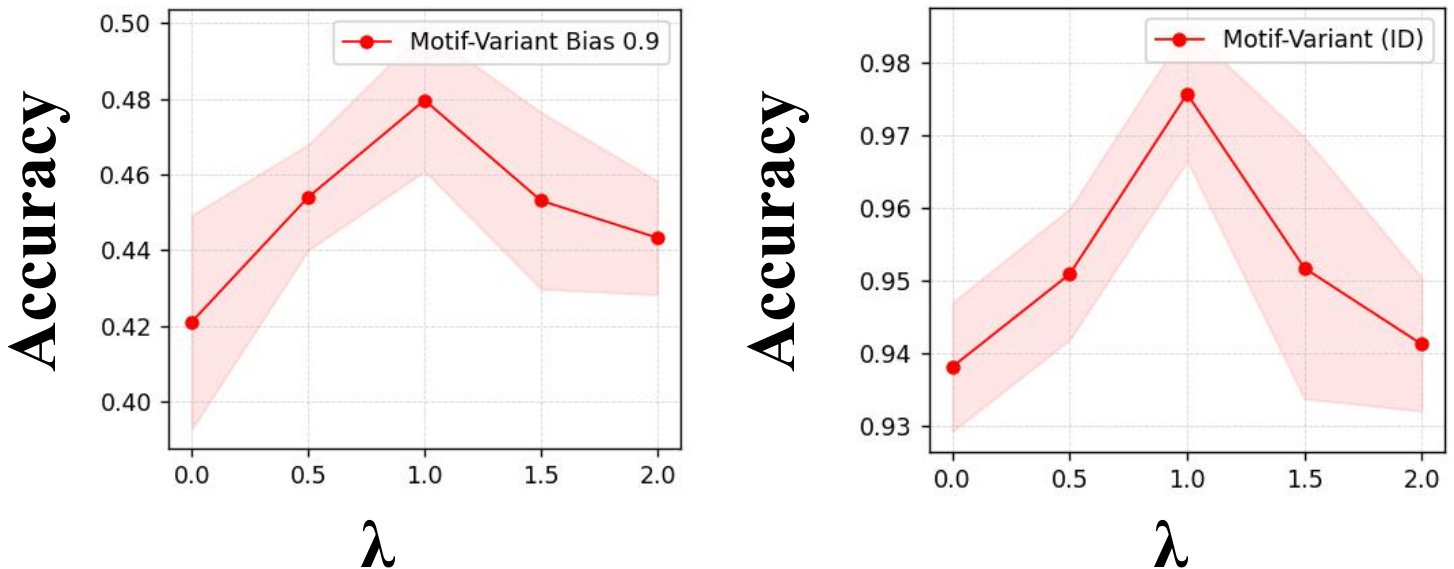} \\
%  		\subcaption{\hspace*{2.3cm}(a) MUTAG}
        \label{pic:mv}
		\end{minipage}
	}\hspace{2mm} 

	\caption{Performance of CLGL on Motif-variant dataset with different hyperparameter values. The translucent region represents the confidence interval of the results.}
	\label{fig:mv}
	%\vskip -0.1in
\end{figure*}

\subsection{Implentation of $f^{P}(\cdot)$}
\label{apx:fp}
Based on the previously determined labels, $f^{P}(\cdot)$ takes the labeled nodes as input, along with the number of neural network layers determined by hyperparameters. Through pooling and multi-Layer perceptron, it ultimately outputs distribution predictions. The prediction is a vector of dimensionality equal to the number of categories, and it has undergone normalization.

\begin{figure*}[htbp]
	\centering
	\subfigure[]{
		\begin{minipage}{0.45\textwidth} 
			\includegraphics[width=\textwidth]{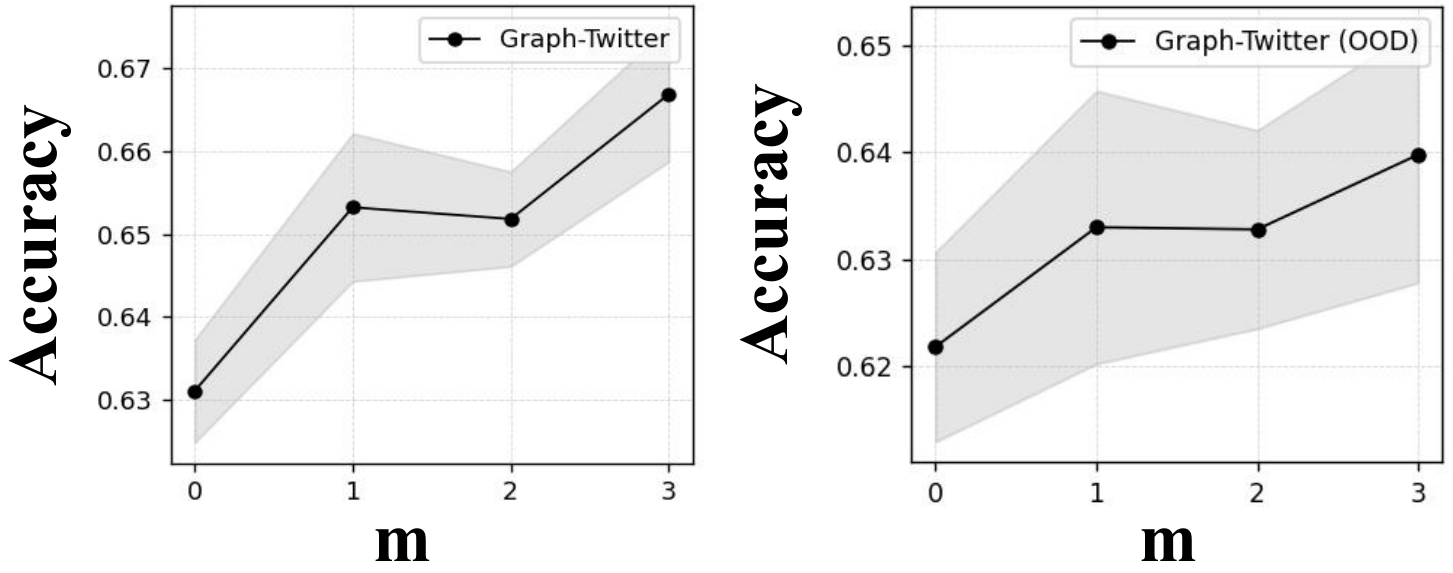} \\
%  		\subcaption{\hspace*{2.3cm}(a) MUTAG}
        \label{pic:m-a}
		\end{minipage}
	}\hspace{2mm} 
	\subfigure[]{
		\begin{minipage}{0.45\textwidth} 
			\includegraphics[width=\textwidth]{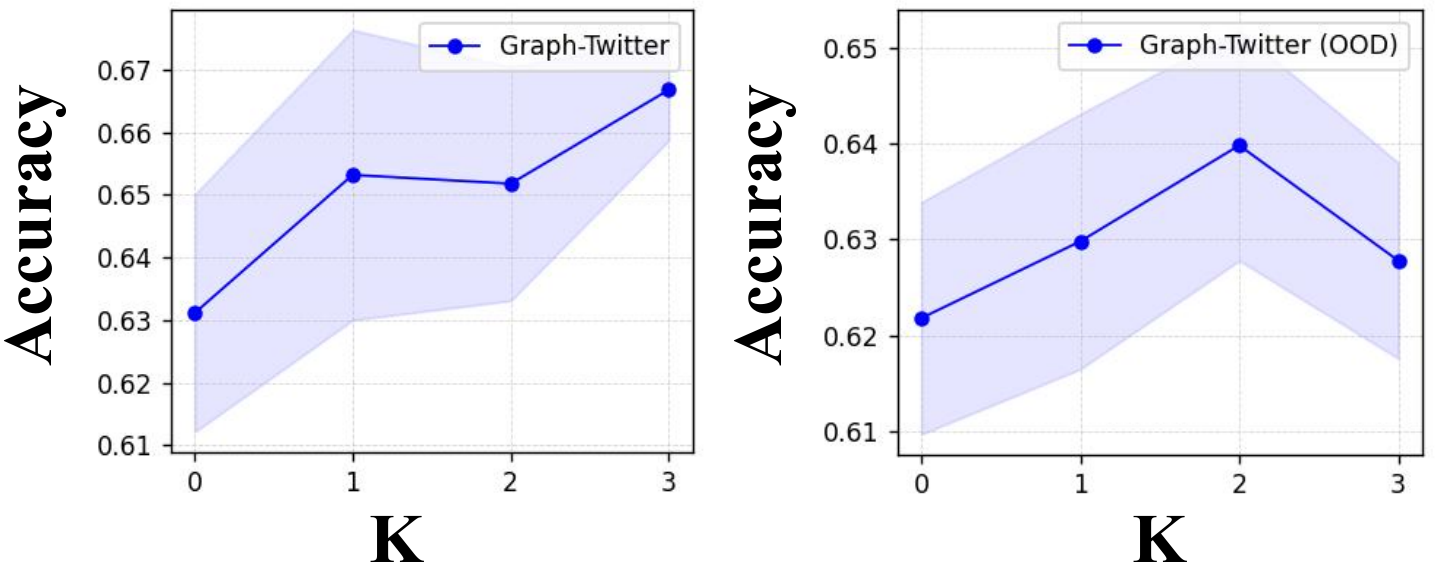} \\
%  		\subcaption{\hspace*{2.3cm}(a) MUTAG}
        \label{pic:m-b}
		\end{minipage}
	}\hspace{2mm} 
 	\subfigure[]{
		\begin{minipage}{0.45\textwidth} 
			\includegraphics[width=\textwidth]{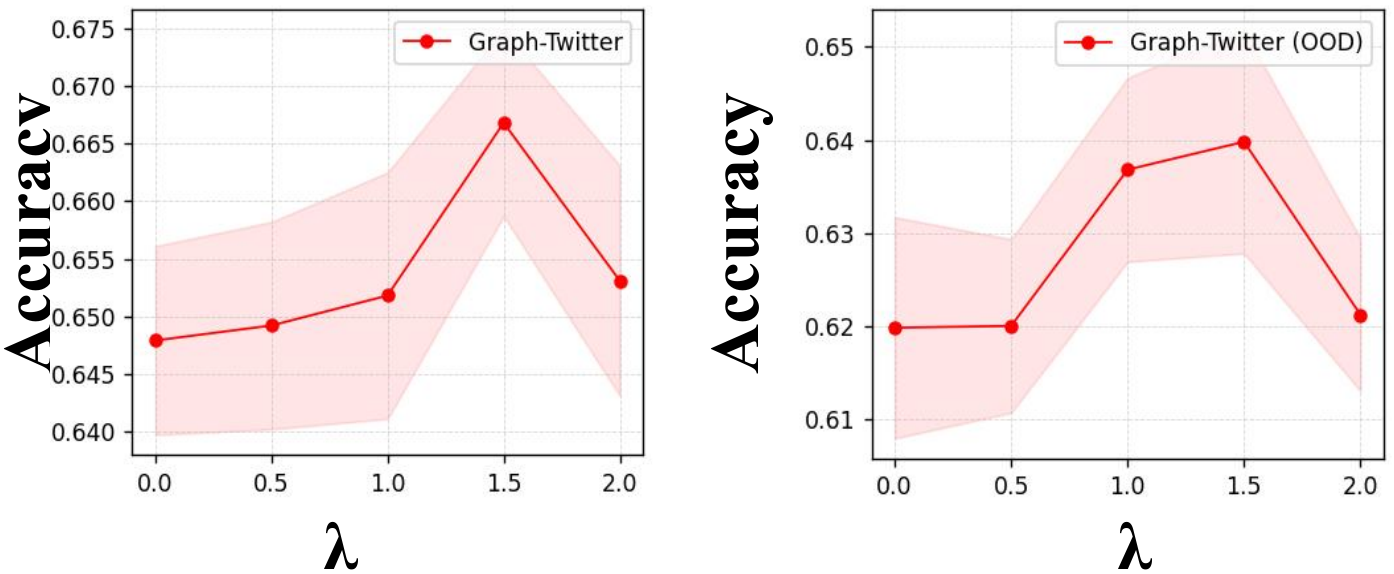} \\
%  		\subcaption{\hspace*{2.3cm}(a) MUTAG}
		\end{minipage}
	}\hspace{2mm} 

	\vskip -0.05in
	\caption{Performance of CLGL on Graph-Twitter dataset with different hyperparameter values. The translucent region represents the confidence interval of the results.}
	\label{fig:tw}
	%\vskip -0.1in
\end{figure*}

% \section{Data Preprocessing}
% \label{apx:preprocess}
% We adopt data preprocessing to mark $\widetilde{X}$ within datasets. For synthetic datasets, including Spurious-Motif and Motif-Variant, as they are artificially generated, we output the marks along with the generation process. Please note that the marks can also be created using approaches such as traversal. For Graph-SST5 and Graph-Twitter Dataset, we write a script to mark the sentence with the word ``but''.

\section{Additional Hyperparameter Experiments}
\begin{figure*}[ht]
	\centering
	\subfigure[]{
		\begin{minipage}{0.45\textwidth} 
			\includegraphics[width=\textwidth]{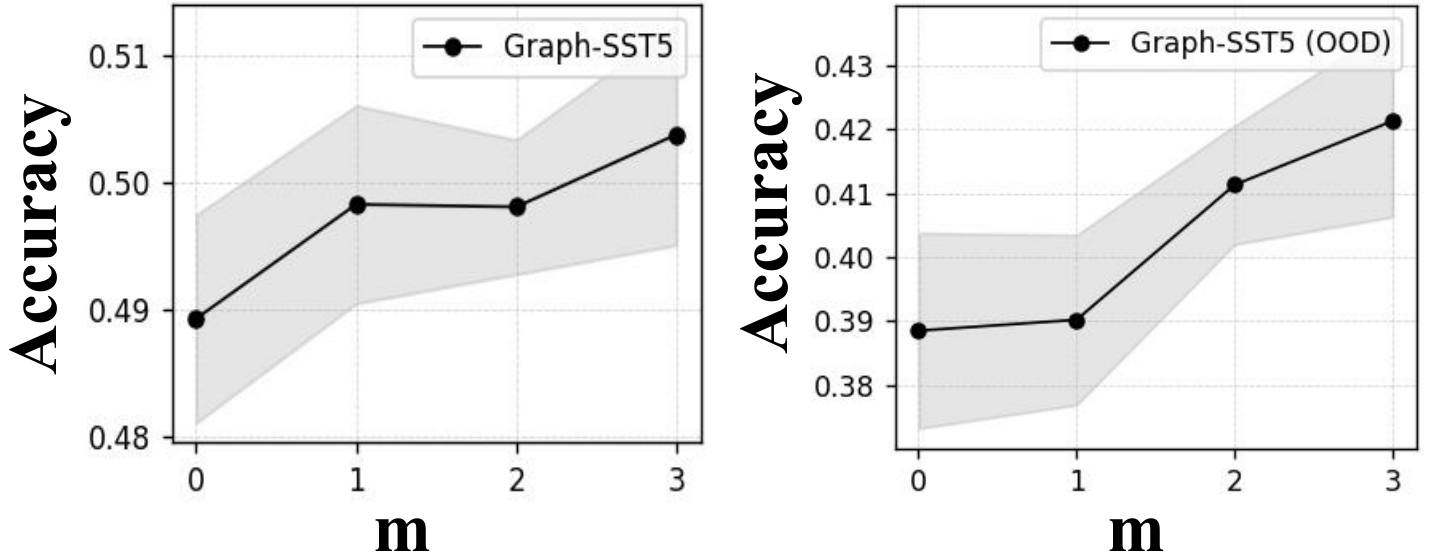} \\
%  		\subcaption{\hspace*{2.3cm}(a) MUTAG}
        \label{pic:m-a}
		\end{minipage}
	}\hspace{2mm} 
	\subfigure[]{
		\begin{minipage}{0.45\textwidth} 
			\includegraphics[width=\textwidth]{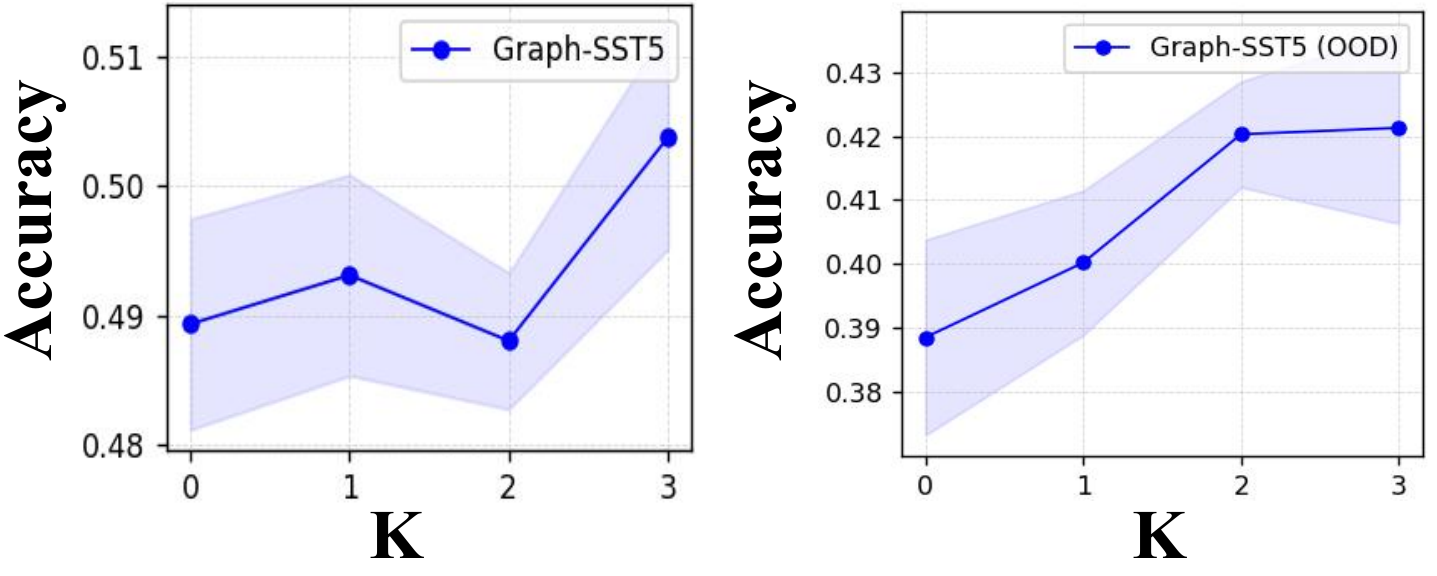} \\
%  		\subcaption{\hspace*{2.3cm}(a) MUTAG}
        \label{pic:m-b}
		\end{minipage}
	}\hspace{2mm} 
 	\subfigure[]{
		\begin{minipage}{0.45\textwidth} 
			\includegraphics[width=\textwidth]{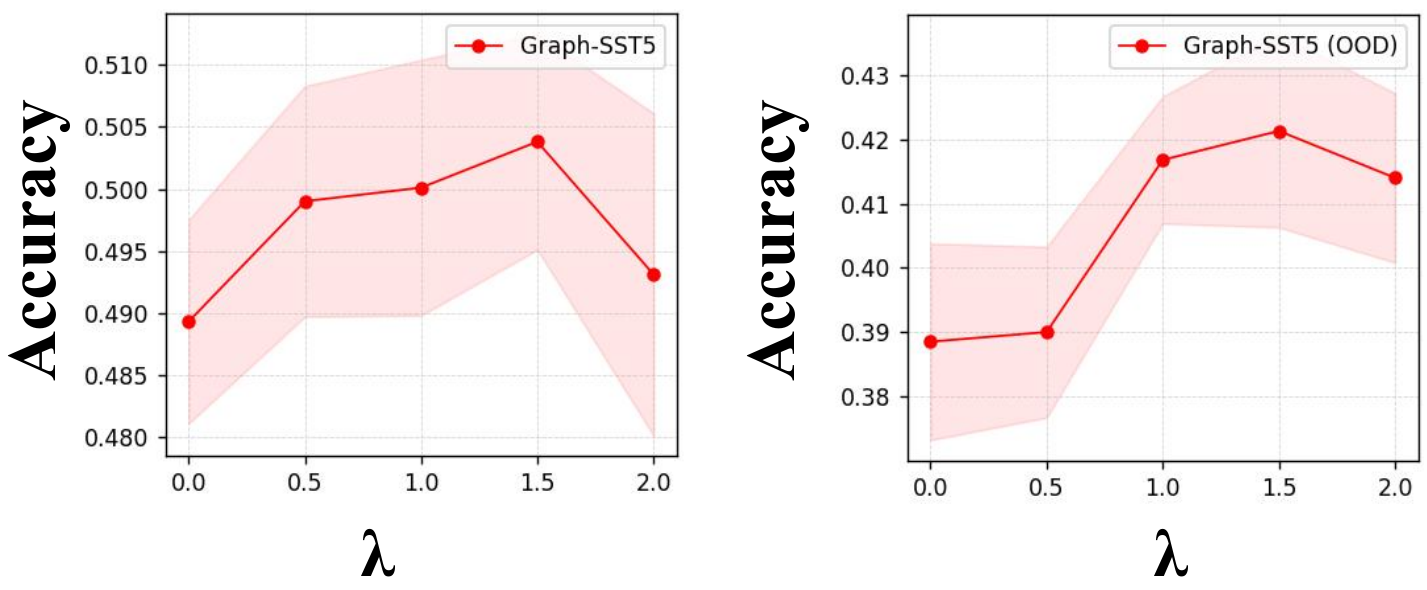} \\
%  		\subcaption{\hspace*{2.3cm}(a) MUTAG}
        \label{pic:m-c}
		\end{minipage}
	}\hspace{2mm} 

	\vskip -0.05in
	\caption{Performance of CLGL on Graph-SST5 dataset with different hyperparameter values. The translucent region represents the confidence interval of the results.}
	\label{fig:5}
	%\vskip -0.1in
\end{figure*}

This section presents further parameter experiments. The experimental results are shown in Figures \ref{fig:spm}, \ref{fig:mv}, \ref{fig:tw} and \ref{fig:5}. From the figures, we can observe that while the optimal values of certain parameters vary across different versions of the datasets, we still chose to use the same parameters to avoid introducing additional prior knowledge.

We can observe from the results that different hyperparameters have a significant impact on the performance of the model. When the hyperparameters are set to zero, indicating minimal influence from our proposed framework on the GNNs, the model's performance deteriorates significantly. This aspect highlights the necessity of the designed structure.
\end{document}